\documentclass[lettersize,journal]{IEEEtran}
\pdfoutput=1
\usepackage{amsmath,amsfonts}
\usepackage{algorithmic}
\usepackage{algorithm}
\usepackage{array}
\usepackage[caption=false,font=normalsize,labelfont=sf,textfont=sf]{subfig}
\usepackage{textcomp}
\usepackage{stfloats}
\usepackage{url}
\usepackage{verbatim}
\usepackage{graphicx}
\usepackage{cite}
\usepackage{tablefootnote}
\usepackage{CJKutf8}

\usepackage{graphicx} 
\usepackage{booktabs}
\usepackage{multibib}
\usepackage{caption}
\usepackage{multirow}
\usepackage[accsupp]{axessibility}
\usepackage{amsthm,amsmath,amssymb}
\usepackage{mathrsfs}
\usepackage{bbding}
\usepackage{url}            %
\usepackage[table]{xcolor}  
\usepackage{balance} 

\definecolor{cvprblue}{rgb}{0.21,0.49,0.74}
\usepackage[pagebackref,breaklinks,colorlinks,citecolor=cvprblue]{hyperref} 

\newlength\savewidth\newcommand\shline{\noalign{\global\savewidth\arrayrulewidth \global\arrayrulewidth 1pt}\hline\noalign{\global\arrayrulewidth\savewidth}}
\newcommand{\tablestyle}[2]{\setlength{\tabcolsep}{#1}\renewcommand{\arraystretch}{#2}\centering\footnotesize}

\makeatletter\renewcommand\paragraph{\@startsection{paragraph}{4}{\z@}
  {.5em \@plus1ex \@minus.2ex}{-.5em}{\normalfont\normalsize\bfseries}}\makeatother

\newcolumntype{x}[1]{>{\centering\arraybackslash}p{#1pt}}
\newcolumntype{y}[1]{>{\raggedright\arraybackslash}p{#1pt}}
\newcolumntype{z}[1]{>{\raggedleft\arraybackslash}p{#1pt}}

\newcommand{\app}{\raise.17ex\hbox{$\scriptstyle\sim$}}

\definecolor{baselinecolor}{gray}{.9}
\newcommand{\baseline}[1]{\cellcolor{baselinecolor}{#1}}

\newcolumntype{*}{>{\global\let\currentrowstyle\relax}}
\newcolumntype{^}{>{\currentrowstyle}}

\definecolor{dt}{gray}{0.7}

\newcommand{\ie}{{\emph{i.e.}}, }

\newcommand{\eg}{{\emph{e.g.}}, }

\usepackage[capitalize]{cleveref}
\crefname{section}{Sec.}{Secs.}
\Crefname{section}{Section}{Sections}
\Crefname{table}{Table}{Tables}
\crefname{table}{Tab.}{Tabs.}

\usepackage{orcidlink}

\newcolumntype{S}{@{}>{\lrbox0}l<{\endlrbox}}  %
\definecolor{lightgreen}{HTML}{D8ECD1}
\newcommand{\better}[1]{\cellcolor{baselinecolor}{#1}}

\begin{document}

\title{Graph Network for Sign Language Tasks}
\author{
Shiwei Gan\textsuperscript{\orcidlink{0000-0003-3360-4321}},~\IEEEmembership{Student Member, IEEE}, 
Yafeng Yin\textsuperscript{\orcidlink{0000-0002-9497-6244}},~\IEEEmembership{Member, IEEE}, 
Zhiwei Jiang\textsuperscript{\orcidlink{0000-0001-5243-4992}},~\IEEEmembership{Member, IEEE}, 
Hongkai Wen\textsuperscript{\orcidlink{0000-0003-1159-090X }},~\IEEEmembership{Senior Member, IEEE}, 
Lei Xie\textsuperscript{\orcidlink{0000-0002-2994-6743}},~\IEEEmembership{Senior Member, IEEE} and 
Sanglu Lu\textsuperscript{\orcidlink{0000-0003-1467-4519}},~\IEEEmembership{Member,~IEEE}

\IEEEcompsocitemizethanks{
\IEEEcompsocthanksitem Shiwei Gan, Yafeng Yin, Zhiwei Jiang, Lei Xie and  Sanglu Lu are with  State Key Laboratory for Novel Software Technology, Nanjing University, Nanjing 210023, China. E-mail: sw@smail.nju.edu.cn, \{yafeng, jzw, lxie, sanglu\}@nju.edu.cn.
\IEEEcompsocthanksitem (Corresponding author: Yafeng Yin)
\IEEEcompsocthanksitem Hongkai Wen is with the Department of Computer Science, University of Warwick, CV4 7AL Coventry, U.K. E-mail: hongkai.wen@warwick.ac.uk.
}
}

 

\IEEEpubidadjcol

\maketitle

\begin{abstract}
Recent advances in sign language research have benefited from CNN-based backbones, which are primarily transferred from traditional computer vision tasks (\eg object identification, image recognition). However, these CNN-based backbones usually excel at extracting features like contours and texture, but may struggle with capturing sign-related features. In fact, sign language tasks require focusing on sign-related regions, including the collaboration between different regions (\eg left hand region and right hand region) and the effective content in a single region.  
To capture such region-related features, we introduce MixSignGraph, which represents sign sequences as a group of mixed graphs and designs the following three graph modules for feature extraction, \ie  Local Sign Graph (LSG) module, Temporal Sign Graph (TSG) module and Hierarchical Sign Graph (HSG) module. 
Specifically, the LSG module learns the correlation of intra-frame cross-region features within one frame, \ie focusing on spatial features. 
The TSG module tracks the interaction of inter-frame cross-region features among adjacent frames, \ie focusing on temporal features.  The HSG module aggregates the same-region features from different-granularity feature maps of a frame, \ie focusing on hierarchical features.
In addition, to further improve the performance of sign language tasks without gloss annotations, we propose a simple yet counter-intuitive Text-driven CTC Pre-training (TCP) method, which generates pseudo gloss labels from text labels for model pre-training. Extensive experiments conducted on current five public sign language datasets demonstrate the superior performance of the proposed model. Notably, our model surpasses the SOTA models on multiple sign language tasks across several datasets, without relying on any additional cues. 
\end{abstract}


\section{Introduction} 
Advancements in computer vision (CV) and natural language processing (NLP) technologies have significantly facilitated the development of sign language (SL) research, including Sign Language Recognition (SLR) and Sign Language Translation (SLT). The former SLR task encompasses Isolated SLR (ISLR)~\cite{hu2021hand} and Continuous SLR (CSLR)~\cite{min2021visual, wei2023improving, gan2024signgraph}, aiming to recognize isolated or continuous signs as corresponding glosses or gloss sequences. While the latter SLT task focuses on translating continuous signs into spoken language~\cite{cihan2018neural}.
In the current neural network-based SL processing framework, the standard approach first uses 2D or 3D CNN-based backbones to extract visual features~\cite{gan23contrastive,chen2022two, tarres2023sign}, then employs 1D CNNs, LSTMs, or Transformers~\cite{cihan2018neural, gan2021skeleton, zhou2021spatial} as temporal modules to capture dynamic changes in sign frames, and finally adopts a CTC decoder to obtain the gloss sequence or a translation model to generate the spoken sentence.

\begin{figure}[t]
	\centering	\includegraphics[width=\columnwidth]{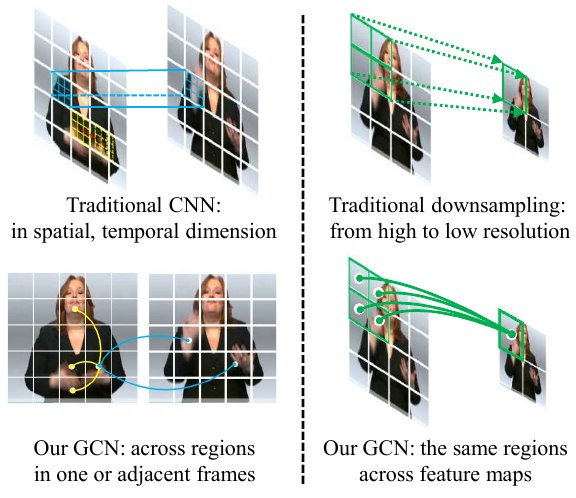}  
	\caption{(Left) Comparison of feature extraction between CNNs and our LSG and TSG.  (Right) Message passing of the same regions among different-granularity/resolution feature maps between downsampling in CNNs and our HSG. }
	\label{fig:intro} 
    \vspace{-2mm}
\end{figure}

The CNN-based backbone has demonstrated its excellent capability for capturing contour-based and texture-based representations in various CV tasks~\cite{geirhos2019imagenet}. 
However, unlike other CV tasks where contour and texture representations are crucial, SL tasks need to focus on both manual and non-manual features, particularly the collaboration of these cues in different regions~\cite{zhou2020spatial, gan2023towards}, 
as shown in the left part of Figure~\ref{fig:intro}. 
Unfortunately, the traditional 2D CNNs (focusing on spatial information) may fail to capture the collaboration of signs across different regions.  
Besides, the commonly-adopted temporal models in SL tasks can also hardly capture cross-region features, since they either focus on the same region or the whole frame, \ie\ NOT cross regions. For example, 1D CNNs can only focus on the same regions in adjacent frames, while LSTM and Transformer layers treat a sign frame as a whole and learn the overall variance of entire frames.
In addition, as shown in the right part of Figure~\ref{fig:intro}, the widely-used CNNs tackle the issue of local-to-global feature extraction by down-sampling feature maps, and adopt the last global features as final extracted features, while ignoring the interaction of the same regions in different granularity feature maps. 
In fact, both the dynamic movements and specific gesture of the body, hands and face, which are reflected across different regions or in a single region, are primary elements for identifying a sign in sign language tasks. 
When moving to the specific sign language tasks, including CSLR and SLT, gloss annotations are often expected. Apparently, in CSLR tasks, gloss annotations are indispensable. However, in SLT tasks, many current SOTA models also rely on gloss annotations to pre-train their backbones with CTC constraints, aiming to improve the SLT performance. When gloss annotations are missing, \ie in gloss-free SLT tasks, how to improve the SLT performance is still a challenging task.

To enhance the CNN-based backbone's ability of extracting sign-related features, existing work has focused on designing various CNN-based backbones by incorporating domain knowledge, such as skeleton information~\cite{gan2021skeleton, wang2020deep}, depth images~\cite{tang2021graph}, or local areas~\cite{zhou2021spatial}. Additionally, some studies have trained their backbones with more training samples through back translation~\cite{zhou2021improving}, cross-modality augmentation~\cite{pu2020boosting}, and pre-trained models~\cite{chen2022simple}. Others have added extra constraints to backbones using methods such as knowledge distillation~\cite{hao2021self} and contrastive learning~\cite{gan23contrastive}. 
In regard to improving the gloss-free SLT performance, the existing models have attempted to introduce attention mechanisms~\cite{yin2023gloss} or contrastive language-image pre-training ~\cite{zhou2023gloss}.

Different from previous work, we build a simple yet effective sign graph neural network (MixSignGraph for short) for SL tasks (including CSLR and SLT), 
in which we take sign frames as graphs of nodes and dynamically build graphs to learn cross-region and one-region features. 
The main intuitions are:  
(1) Sign-related features involve the correlation of different regions in one frame, such as face region and hand regions. Thus, we propose a local sign graph $LSG$ module by dynamically building graphs based on nodes in each frame, to learn intra-frame cross-region features. 
(2) The dynamic movements in the body, hands, and face are reflected in different regions among adjacent frames. Thus, we propose a temporal sign graph $TSG$ module by dynamically connecting sign-related regions based on nodes in adjacent frames, to learn inter-frame cross-region features. 
(3) The content in a single region can be represented by different granularity feature maps generated by down-sampling. Thus, we propose a hierarchical sign graph $HSG$ module by connecting the same regions of different-granularity feature maps, to learn hierarchical one-region features.
Besides, to further improve the performance of sign language tasks without gloss annotations, we propose a simple yet counter-intuitive Text-driven CTC Pre-training (TCP) method, which generates pseudo gloss labels from text labels for model pre-training. We make the following contributions:

\begin{itemize}
    \item  Different from CNN-based SL models, we propose a graph-based SL model (MixSignGraph), which processes sign frames as graphs of nodes and builds three graph modules: a local sign graph module $LSG$ that learns the correlation of cross-region features in one frame, a temporal sign graph module $TSG$ that learns the interaction of cross-region features among adjacent frames, and a hierarchical sign graph $HSG$ module that aggregates the same-region features from different-granularity
feature maps of a frame.

    \item A simple, effective, yet counter-intuitive training method named Text-driven CTC Pre-training (TCP) is proposed for gloss-free SLT task, which generates pseudo gloss annotations based on text labels.
        
    \item The extensive experiments on common sign language tasks (including CSLR and SLT) over five public datasets demonstrate the superiority of the proposed model MixSignGraph, which outperforms SOTA models and does not use any extra cues.
    
\end{itemize}
 
This work is an extension of the conference paper~\cite{gan2024signgraph}
with improvements in the following aspects. (1) We propose a new module named Hierarchical Sign Graph ($HSG$) module that aggregates the same-region features from different-granularity feature maps of a frame, and the proposed $HSG$ module further brings a notable performance gain.  (2) We propose a new simple yet effective pre-training method for gloss-free SLT tasks, which significantly improves the performance of gloss-free SLT and bridges the translation performance gap between gloss-free and gloss-based SLT models. 
(3) We extend the original framework to three more
SLT downstream tasks (including Sign2Text, Sign2Gloss2Text and Gloss-free Sign2Text), and adopt two larger public datasets (\ie How2Sign~\cite{duarte2021how2sign} and OpenASL~\cite{shi2022open}) 
for evaluation. Our newly designed framework achieves
SOTA performances on most of the downstream tasks across several datasets.

\section{Related Work}

In this section, we first give a literature review for video-based
sign language tasks. Then we present an
overview about the usage of graph convolutional networks in SL tasks.  

\subsection{Sign Language Tasks} 
We focus on two common sign language tasks, \ie Continuous Sign Language Recognition (CSLR) and Sign Language Translation (SLT).
The goal of CSLR is to interpret continuous sign sequences as corresponding gloss sequences~\cite{koller2017re}, while SLT aims to translate the sign language into spoken language~\cite{gan2023towards}.

The unique grammatical rules of sign language lead to some differences between the frameworks of CSLR and SLT tasks. Nevertheless, both frameworks contain a visual backbone to capture visual features from videos. 
However, it is a challenging task for the widely adopted CNN-based backbones to focus on sign-related features, since CNN backbones are initially designed to capture the texture and contours of objects~\cite{geirhos2019imagenet}. 
In fact, in SL tasks, identifying a sign needs to focus on the feature of a single region, and the cross-region features both in one frame (\ie the correlation of  hands and face regions) and among adjacent frames (\eg hand motions, facial expression changes). 
To guide CNN-based backbones extract sign-related features, some current SL models introduce prior knowledge (\eg local areas~\cite{camgoz2020multi, ren2016faster}, skeletons~\cite{gan2021skeleton, zuo2022c2slr, chen2022two} and depth images~\cite{tang2021graph}) to manually inject expert knowledge into the backbone, making it focus on SL-related areas. While these models usually rely on extra algorithms (\eg Openpose~\cite{cao2018openpose} or HRNet~\cite{sun2019deep}) or tools (\eg depth camera) to obtain skeleton data or depth images. 
Some other SL models try to add additional constraints (\eg knowledge distillation~\cite{hao2021self,min2021visual} and contrastive learning~\cite{gan23contrastive}) to optimize their backbones. Besides, there are also some SL models that turn their attention to obtaining more sign language samples by back translation~\cite{zhou2021improving}, cross modality augmentation~\cite{pu2020boosting}, pre-training with other sign language related datasets~\cite{chen2022simple, wei2023improving} and adopting pre-trained visual models as their backbones~\cite{wong2024sign2gpt}.  

In regard to the differences between frameworks in SL tasks, CSLR models adopt CTC module~\cite{koller2019weakly} to get recognized gloss sequences. While SLT~\cite{cihan2018neural} is usually limited to seq2seq architecture and often adopts a RNN-like/transformer decoder~\cite{gan23contrastive} or a pre-trained language model (\eg mBART~\cite{liu2020multilingual}, GPT2~\cite{radford2019language}) to get the translated sentence. 
In these SLT models, the traditional training paradigm usually involves two stages: first pre-training their backbone using gloss sequences (equivalent to the CSLR task), then connecting a pre-trained translation model (\eg mBART~\cite{liu2020multilingual}) and fine-tuning the entire model. 
In the first stage, using labelled gloss sequence for pre-training is critical for improving SLT performances, since this pre-training can assist the model in learning segmentation and semantic information of sign language.    
However, annotating gloss sequences for sign language videos is labor-intensive and cumbersome. Therefore, gloss-free SLT tasks have gained attention~\cite{yin2023gloss, zhou2023gloss}. 
The existing gloss-free SLT models usually adopt attention mechanism~\cite{yin2023gloss} or contrastive language-image pre-training~\cite{zhou2023gloss} to improve SLT performances. 
However, these models still exhibit a significant performance gap when compared with gloss-based SLT models.

\begin{figure*}[t]
	\centering
	\includegraphics[width=\textwidth]{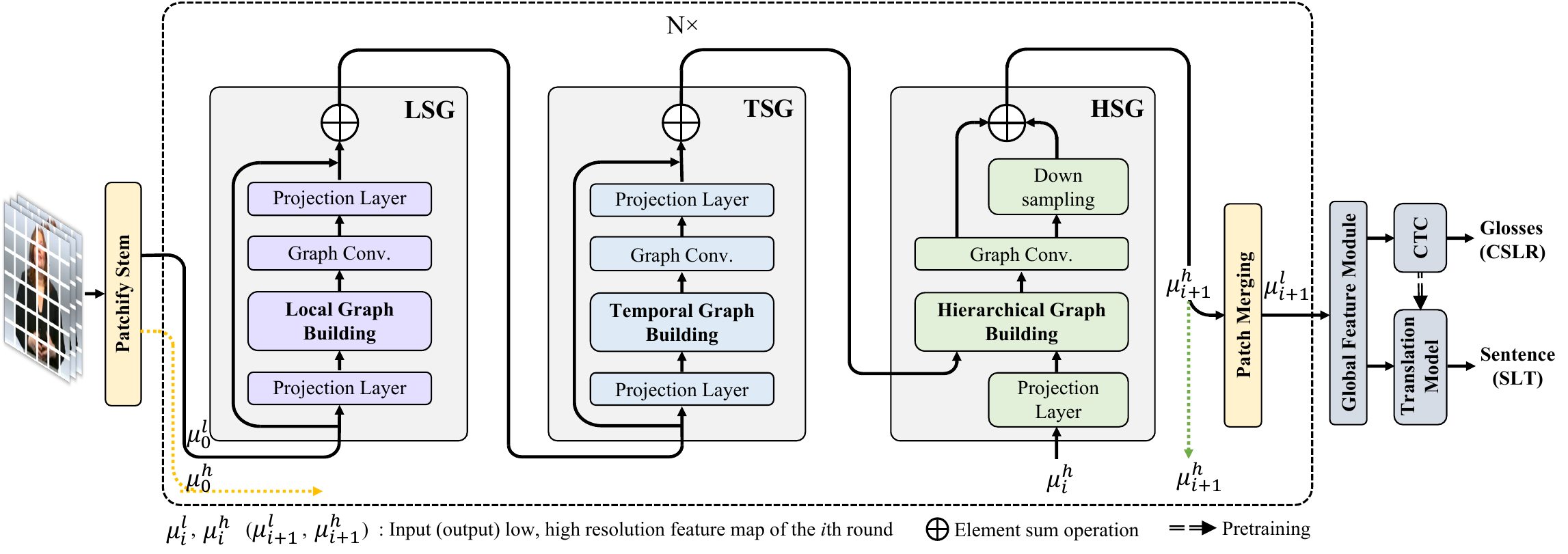} 
  \vspace{-2mm} 
	\caption{The proposed SignGraph architecture. The order of LSG, TSG, HSG is just an example, and the order can be changed.}
	\label{fig:signgraph}
 \vspace{-3mm}
\end{figure*}

\subsection{Graph Convolutional Network}  
Due to the graph structures, graph convolution layers ~\cite{kipf2016semi} are more flexible for message passing and aggregation, when compared with CNN, RNN-like or transformer layers. 
Considering the requirement of graphs, previous work has primarily focused on applying graph convolution networks (GCN) to real-world structured data, \eg social network datasets~\cite{cai2018simple, Yanardag} and bioinformatics datasets~\cite{debnath1991structure}. 
After that, GCN was also applied to skeleton data, where the connections between joints render the skeleton data inherently suitable for GCN. With the skeleton data, GCN has been extensively applied to action recognition tasks \cite{stgcn2018aaai}, and then extended to SL tasks~\cite{gan2021skeleton, kan2022sign, hu2023signbert+}. 
In SL tasks, some models utilized GCNs to guide CNN-based backbones to extract skeleton-related features~\cite{gan2021skeleton, parelli2022spatio}. 
While others ignored RGB features and applied GCN to explore sign features directly with skeletons~\cite{kan2022sign,jiao2023cosign}. 
In the above models, applying graph convolution networks often relied on structured data (\eg skeleton). 
To extract efficient features while reducing the dependence on structured data, the recent model VIT~\cite{dosovitskiy2020image}, which processes an image as patches with a transformer model, has replaced CNNs as a de-factor architecture in many fields.
After that, the VIG model~\cite{han2022vision} directly represents an image as a graph structure to capture irregular and complex objects, and achieves superior performance on image recognition tasks. 
Inspired by VIG, we build a simple yet effective GCN-based framework that treats sign sequences as graphs of nodes and focuses on sign-related regions, including the collaboration across different regions and the effective content in a single region.

\section{Method}
\subsection{Overall Framework}
For a sign language sequence ${f}$=$\{{f_i}\}_{i=1}^{\theta }$ with ${\theta}$ frames, the target of CSLR task is to get a recognized gloss sequence $g=\{g_i\}_{i=1}^{\vartheta}$  with ${\vartheta }$ glosses, while the target of SLT task is to generate a spoken language sentence $t=\{t_i\}_{i=1}^ { \varsigma }$ with $\varsigma$ words based on input $f$.

As shown in Figure~\ref{fig:signgraph}, our MixSignGraph model consists of three key modules, \ie Local Sign Graph $LSG$ module, Temporal Sign Graph $TSG$ module, and Hierarchical Sign Graph ($HSG$) module.
When given the video frames, a patchify stem is adopted to convert each frame into a set of patches (\ie nodes) $v_i=\{v_{ij}\}_{j=1}^{N}$. 
Then, the three key modules are utilized to construct graphs and capture sign-related features as follows.
First, by connecting $\mathsf{K}$ nearest neighbors for each node in one frame, we build a dynamic local sign graph $G^l_i$ for each frame $f_i$, which is then applied with a graph convolutional layer to learn the correlation of cross-region features within one frame.
Second, by connecting sign-related regions between two adjacent frames, we build a dynamic temporal graph $G^t_i$, which is then applied with a graph convolutional layer to track the interaction of cross-region features among adjacent frames.
Third, by aggregating the same-region features from different feature maps of a frame, we build a hierarchical sign graph $G^h_i$, which is then applied with a graph convolutional layer to bidirectionally exchange features and mine different granularity one-region features. 
In addition, we stack the three modules and adopt patch merging to get larger-size patches. 
After that, following previous CSLR models~\cite{min2021visual, gan23contrastive}, we adopt a global feature module to learn global changes of whole frames. 
Finally, to tackle CSLR tasks, a classifier and a widely-used CTC loss~\cite{graves2006connectionist} are adopted to predict the probability $p(g|f)$ of the target gloss sequence. 
When moving to SLT tasks, an extra translation model is utilized to convert feature sequences or gloss sequences into the spoken language sentence. It is worth mentioning that a Text-driven CTC Pre-training (TCP) mechanism is further proposed for gloss-free SLT task, while considering the high cost of gloss annotations. 

\begin{figure}[t]
	\centering
            \includegraphics[width=\columnwidth]{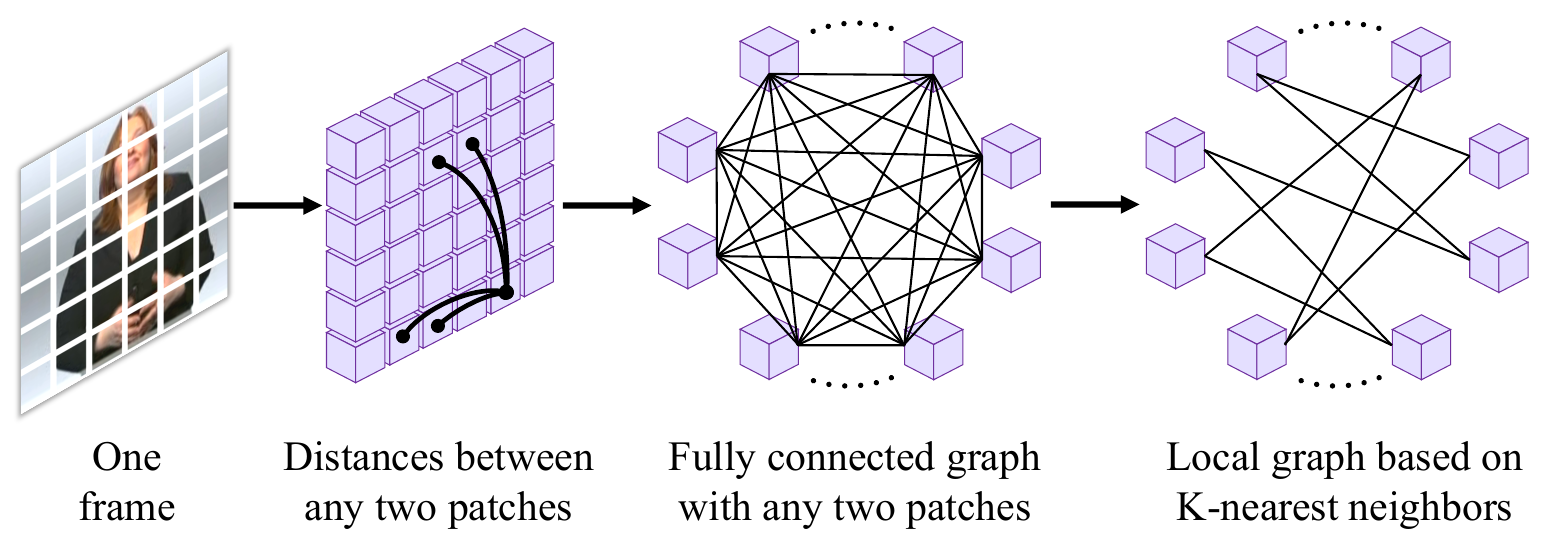}
            \caption{Graph construction in local sign graph module.}
	\label{fig:localG} 
\end{figure}

\subsection{Sign Graph Learning}

\subsubsection{Patchify Stem}  In order to apply GCNs directly to sign frames instead of skeleton data, we need to convert RGB data into nodes. Following previous work~\cite{dosovitskiy2020image, wu2021cvt}, for each frame $f_i\in \mathbb{R}^{[H\times W \times 3]}$ with height $H$ and width $W$, we first convert it into a number of patches (or nodes) $v_i=\{v_{ij}\}_{j=1}^{N}$ with a patchify stem $\mathcal{PS}$.
\begin{equation}
	\begin{aligned}
		 \mu_i = \{\mu_{ij}\}_{j=1}^{N} = \mathcal{PS}(f_i)
	\end{aligned}
\end{equation} Here, $v_{ij}$ represents the $j$th node in the $i$th frame $f_i$, $\mu_{ij}\in \mathbb{R}^{D}$ denotes node features of $v_{ij}$, $D$ is the feature dimension, $N$$=$$HW/P^2$ is the number of patches, and $P$ is the patch size.

\subsubsection{Local Sign Graph Learning} For local sign features in each frame, SL models need to focus on the correlation between different regions, such as the face and hand regions. 
To effectively learn these intra-frame cross-region features, it is essential to construct a local sign graph that facilitates the interaction between correlated regions. 
As shown in Figure~\ref{fig:signgraph} and \ref{fig:localG}, to construct such a graph, we initially use a projection layer with weights $\Theta^l_1$ to map node features $\mu_i$ to the space where the distance function is applied, resulting in the transformed features $\mu'_i = \mu_i \Theta_1^l$. Then, we adopt $K$-Nearest Neighbors (KNN) algorithm to find top $\mathsf{K}_l$ nearest neighbors $\mathcal{N}_{\mathsf{K}_l}(v_{ij})$ for each node based on distance function $\mathcal{DIS}$~\cite{han2022vision}.
\begin{equation}
	\begin{aligned}
    \mathcal{N}_{\mathsf{K}_{l}}(v_{ij}) =\{v_{ik} | v_{ik} \in  TopK (\{\mathcal{DIS}(v_{ij}, v_{ik})\}_{k=1}^{N}) \}
	\end{aligned}
\end{equation}
Here, $TopK$ function outputs the top $K$ nearest neighbors for $v_{ij}$ based on node distances. 
After that, we add an undirected edge $e(v_{ij}, v_{ik})$ between $v_{ij}$ and $v_{ik}$, where $v_{ik} \in \mathcal{N}_{\mathsf{K}_{l}}(v_{ij})$. Consequently, for frame $f_i$, we obtain the edge set $e_i^l = \{ e(v_{ij}, v_{ik}) \mid j \in [0, N), v_{ik} \in \mathcal{N}_{\mathsf{K}_{l}}(v_{ij}) \}$, and construct a local graph $G^L_i = \{ v_i, e_i^l \}$. Following this, we employ a graph convolutional layer $\mathcal{GCN}_l$ and another projection layer with weights $\Theta_2^l$ to aggregate cross-region features for each frame.
\begin{equation}
	\begin{aligned}
		 \mu_i = \mu_i +   \mathcal{GCN}_L(\mu'_i, e_i^l) \Theta_2^l \\  
	\end{aligned}
\end{equation}

Unlike VIG~\cite{han2022vision}, the $LSG$ module does not contain extra modules (\eg feed-forward network), while only containing one graph convolutional layer and two projection layers. The later experiments will show that our model can achieve a convincing performance even with this simple design.

\begin{figure}[t]
	\centering
            \includegraphics[width=\columnwidth]{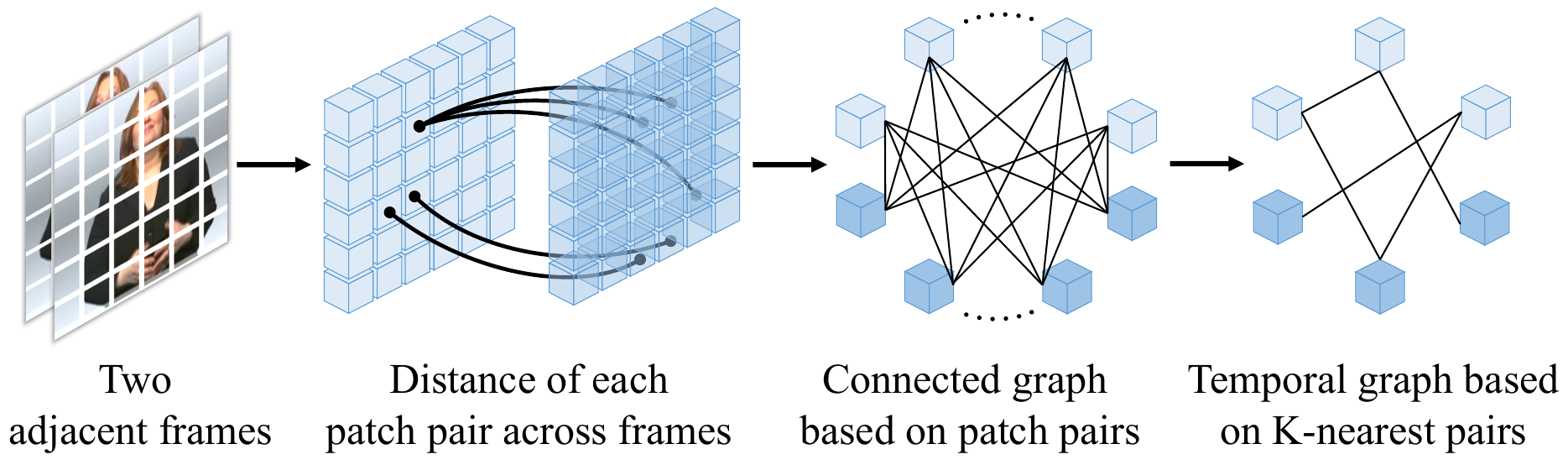}
            \caption{Graph construction in temporal sign graph module.}
	\label{fig:temporalG} 
\end{figure}   

\subsubsection{Temporal Sign Graph Learning} 
To capture temporal sign features, sign language models need to emphasize the dynamic movements of the body, hands and face, which are key to identifying signs and often span different regions across consecutive frames. To achieve this goal, we design a Temporal Sign Graph ($TSG$) module that dynamically creates connections between regions in consecutive frames and learns inter-frame cross-region features. 

As shown in Figure~\ref{fig:signgraph} and~\ref{fig:temporalG}, for two adjacent frames $f_i$ and $f_{i+1}$, which contains $2N$ nodes $\{v_{ij}\}_{j=1}^{N}$ and $\{v_{(i+1)k}\}_{k=1}^{N}$, we first apply a projection layer with weight $\Theta_1^t$ to map node features $\mu_i$ to $\mu''_i = \mu_i \Theta_1^t$. 
Next, we compute a $N \times N$ distance matrix $M$ by calculating distances between $v_{ij}$ and $v_{(i+1)k}$, where $\{M[j,k] = \mathcal{DIS}(v_{ij}, v_{(i+1)k})| j \in [0,N), k \in [0,N)\}$. 
With the matrix $M$, the $TSG$ module selects the top $\mathsf{K}_t$ node pairs between two adjacent frames and adds undirected edges for each pair. 
In this way, we obtain the temporal edge set $e^t_{i}$ between frames $f_i$ and $f_{i+1}$, and construct the corresponding temporal graph $G^{T}_{i}=\{(v_i, v_{i+1}), e^t_{i}\}$.
\begin{equation}
	\begin{aligned}  
		e^t_{i}=\{e(v_{ij},v_{(i+1)k}) |  \mathcal{DIS}(v_{ij}, v_{(i+1)k} ) \in TopK(M)\}
	\end{aligned}
\end{equation}
Finally, we combine all the edges and nodes, and construct a whole temporal graph for the entire video. Besides, a graph convolutional layer $\mathcal{GCN}_t$ and another projection layer with weights $\Theta_2^t$ are applied to the whole temporal graph, to aggregate cross-region features from adjacent frames.
 \begin{equation}
	\begin{aligned}
		 \{\mu_i\}^{\theta}_{i=1} = \{\mu_i\}^{\theta}_{i=1} + \mathcal{GCN}_t\left ( \{\mu''_i\}^{\theta}_{i=1}, 
   \{e_i^t\}^{\theta-1}_{i=1} \right)  \Theta_2^t 
	\end{aligned}
\end{equation} 

Unlike the $LSG$ module, which selects the top $\mathsf{K}_l$ nearest nodes for each node $v_{ij}$ to build a dense graph, the $TSG$ module selects limited node pairs and constructs a sparse graph, aiming to focus on dynamic changes in key areas rather than background areas.  
In addition, constructing a sparse graph in the $TSG$ module is possible to prevent the over-smoothing phenomenon~\cite{huang2020tackling}, which may reduce the distinctiveness of node features.

\begin{figure}[t]
\centering
\includegraphics[width=0.75\columnwidth]{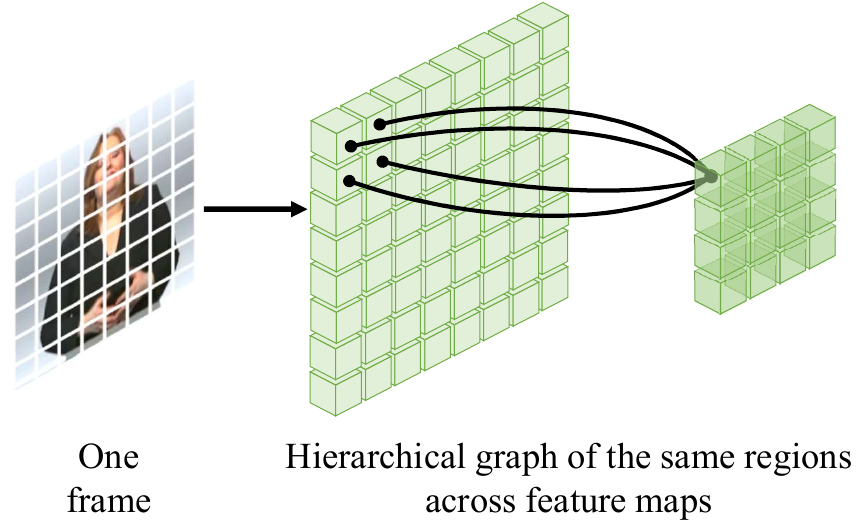}
            \caption{Graph construction in hierarchical sign graph module.} 
	\label{fig:HSG} 
\end{figure}

\subsubsection{Hierarchical Sign Graph Learning} 
When using CNNs to extract features, the content of a region can be represented with different granularity feature maps by downsampling. However, these traditional CNNs can only downsample one region from local to global representations (\ie unidirectional feature extraction), while not providing the interaction between different granularity features (\ie NOT bidirectional feature interaction).
This means that traditional CNNs do not utilize the final global feature map to guide the previous local feature extraction in return, and only adopt the final global feature map to represent a region, while ignoring the previous feature maps.
To mine richer features of a region, we design a Hierarchical Sign Graph ($HSG$) module, which connects the same regions in feature maps of a frame and enables bidirectional exchanges of different granularity feature maps.     
In this way, we can combine hierarchical features of the same region, and make the model focus on both global shapes and detailed local features of a region at the same time.


As shown in Figure~\ref{fig:HSG}, given two feature maps $\mu_i^h \in \mathbb{R}^{[H^h \times W^h \times C^h]}$ in high resolution and $\mu_i^l \in \mathbb{R}^{[H^l \times W^l \times C^l]}$ in low resolution for frame $f_i$, we have $N^h = H^h W^h$ nodes $v_i^h$ for $\mu_i^h$ and $N^l = H^l W^l$ nodes $v_i^l$ for $\mu_i^l$.
Here, $H^h$, $W^h$, $C^h$, $H^l$, $W^l$ and $C^l$ are the height, width and the number of channels of $\mu^h$ and $\mu^l$ respectively, and $\frac{H^h}{H^l}=\frac{W^h}{W^l}=s>0$ where $s$ is the downsampling factor. 
As shown in Figure~\ref{fig:signgraph} and \ref{fig:HSG}, to construct the hierarchical sign graph, we first use a projection layer with weights $\Theta_1^h$ to map $\mu_i^h \in \mathbb{R}^{[H^h \times W^h \times C^h]}$ to $\mu_i^{h'} = \mu_i^h \Theta_1^h$, where $\mu_i^{h'} \in \mathbb{R}^{[H^h \times W^h \times C^l]}$. Then, we add an undirected edge $e(v_{ij}^h, v_{ik}^l)$, if the node $v_{ij}^h$ in $\mu_{i}^{h'}$ and the node $v_{ik}^l$ in $\mu_{i}^l$ correspond to the one region. 
\begin{equation}
	\begin{aligned}  
e_i^b &= \{ e(v_{ij}^h, v_{ik}^l) \mid j \in [0, N^h), \\
k &= \left( \left\lfloor \frac{\left\lfloor \frac{j}{W^h} \right\rfloor}{s} \right\rfloor \times W^l + \left\lfloor \frac{j \% W^h}{s} \right\rfloor \right)
	\}
	\end{aligned}
\end{equation}
In this way, we can get the edge set $e^b_{i}$ between $\mu_i^l$ and $\mu_i^h$ corresponding to the frame $f_i$, and obtain the hierarchical sign graph $G^{H}_{i} = \{(v_i^{h'}, v_i^l), e^b_{i}\}$. 
After that, we apply a graph convolutional layer $\mathcal{GCN}_H$ to enable bidirectional exchanges of the same regions in different-granularity feature maps. Finally, we use a convolutional layer with weights $\Theta_2^b$ and a stride $s$ to downsample $\mu_i^h$ for the final feature fusion.
\begin{equation}
	\begin{aligned}  
\mu_{i}^{h}, \mu_{i}^{l} &= \mathcal{GCN}_{H}(\{\mu_{i}^{h'}, \mu_{i}^{l}\}, e_{b}^{i})  
	\end{aligned} 
\end{equation}
\begin{equation}
	\begin{aligned}  
\mu_i^l =\mu_i^l +\mu_i^{h'}  \Theta^b_2 
	\end{aligned}
\end{equation}

Unlike $LSG$ and $TSG$, which dynamically build edges based on the KNN algorithm to capture cross-region features, the $HSG$ focuses on fusing features belonging to the same region at different granularities, aiming to extract richer features from a region. 

\subsection{Mix Sign Graph Convolutional Network}  
Considering that patches with a fixed window may not effectively capture sign features (\eg hand regions may be split by different patches), we construct our model MixSignGraph with a hierarchical architecture~\cite{liu2021swin}, ensuring that node representations in graphs can capture region-level features at different granularities. 

\subsubsection{Patch Merging}
To produce a multiscale representation, we design a patch merging module, which can reduce the number of patches (nodes) and expand the receptive field of patches. 
Specifically, for $N$ patches (or nodes) with a resolution of $H \times W$ in frame $f_i$, the patch merging module applies convolutional layers to downsample feature maps by a factor of 2, to generate larger-size patches. Then, we can get a set of larger-size nodes $\{\Bar{v}_{ij}\}_{j=1}^{N/4}$ for frame $i$, along with their corresponding node features $\{\Bar{\mu}_{ij}\}_{j=1}^{N/4}$, where $\Bar{\mu}_{ij} \in \mathbb{R}^{2D}$.

\subsubsection{Mix Sign Graph Learning}
With the nodes $\{\Bar{v}_{ij}\}_{j=1}^{N/4}$ of a larger size for frame $i$, we adopt the $LSG$, $TSG$ and $HSG$ modules again, aiming to extract sign-related features from different-size regions.   
In this way, there will be more than one $LSG$, $TSG$, $HSG$ module in our model. In regard to the number of $LSG$, $TSG$ and $HSG$ modules, it is determined based on the experimental results described in Section~\ref{sec:ablation}. In this paper, our baseline settings incorporate two $LSG$, $TSG$ and $HSG$ modules in the model MixSignGraph. 
For convenience, we represent the $i$th $LSG$, $TSG$ and  $HSG$ module as $LSG_i$, $TSG_i$ and $HSG_i$, respectively.
The corresponding hyperparameters of $LSG_i$ and $TSG_i$ are $\mathsf{K}_l^i$ and $\mathsf{K}_t^i$. 

For clarity, we name the previously proposed model from the conference paper~\cite{gan2024signgraph} as \textbf{MultiSignGraph} (which includes two $LSG$ and $TSG$ modules) and the current model (incorporating two $LSG$, $TSG$ and $HSG$ modules) as \textbf{MixSignGraph}.

\begin{figure}[t]
	\centering \includegraphics[width=0.85\columnwidth]{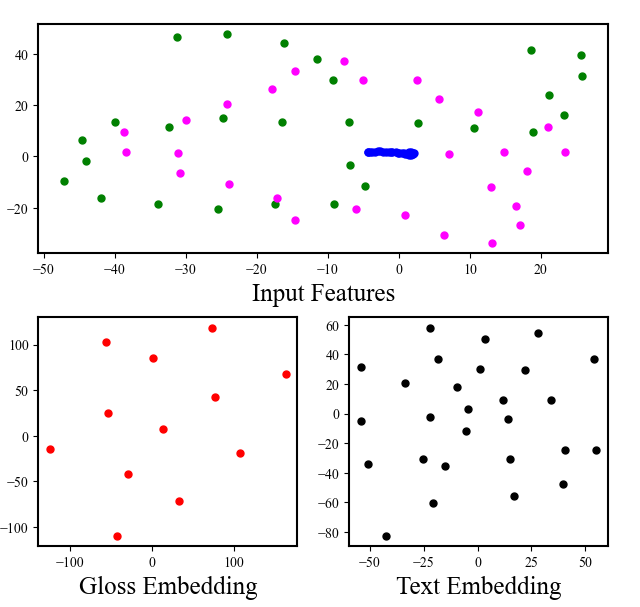} 
 \caption{ Visualization of feature distribution of one test sample in PHOENIX14T dataset, which is displayed via t-SNE~\cite{van2008visualizing}. Blue, green, pink points represent the input features of the translation model by NOT using TCP, using TCP, using CTC pre-training with gloss annotations. Red, black points represent the corresponding gloss sequence, text sequence embedding.}
\label{fig:tcp_vis}
\end{figure} 

\subsection{Text-driven CTC Pre-training}
\subsubsection{Gloss-based Tasks}
The CSLR tasks and gloss-based SLT tasks rely on gloss annotations for model training. For CSLR tasks, CTC loss with gloss annotations has become de-facto loss function in current CSLR models. While in gloss-based SLT tasks, the current SOTA SLT models~\cite{chen2022simple, chen2022two} rely on pre-training, in which CTC loss with gloss sequences is used to optimize the backbone to extract effective video level features, as described in Equation~\ref{equ:ctcloss}. After the pre-training, these SLT models further introduce a pre-trained translation model (\eg mBART~\cite{liu2020multilingual}) and fine-tune the translation model for SLT, as described in Equation~\ref{equ:CEloss}.
 Here, $\mathcal{L}_{CTC}$ denotes CTC loss function, $\mathcal{L}_{CE}$ denotes cross-entropy loss function, $\hat{g}$ and $g$ respectively denote recognized gloss sequence and labeled gloss sequence, $\hat{t}$ and $t$ respectively denote translated sentence and labeled sentence, $\Theta_R$ and $\Theta_T$ denote the parameters of recognition and translation models.
\begin{gather}  
 \label{equ:ctcloss} 
 \min_{\Theta_R} \mathcal{L}_{CTC}(\hat{g}, g)  \\
 \label{equ:CEloss} 
 \min_{\Theta_R, \Theta_T} (\mathcal{L}_{CTC}(\hat{g}, g) +\mathcal{L}_{CE}(\hat{t}, t)) 
\end{gather}

\subsubsection{Gloss-free Tasks}
The above pre-training process enables the backbone to learn segmentation and semantic information from gloss annotations, which are crucial for improving the SLT performance. 
However, in the other SLT tasks, \ie gloss-free SLT tasks, there are not gloss annotations, since annotating glosses is a labor-intensive task. 

\textbf{Text-driven CTC Pre-training (TCP)}: 
To improve the performance of these gloss-free SLT tasks, we propose a simple, effective, yet counter-intuitive training method called Text-driven CTC Pre-training (TCP), which generates pseudo gloss sequences from labeled sentences for pre-training. 
Specifically, in TCP, we obtain the pseudo gloss sequence $t'$ by simple pre-processing labeled sentence $t$, including removing punctuations, lemmatization and word-level tokenization. Then, we directly use the CTC loss with pseudo gloss sequence $t'$ to optimize the backbone during initial pre-training, as described in Equation~\ref{equ:ctcloss1}. After that, we fine-tune the entire model for translation, as depicted in Equation~\ref{equ:CEloss1}. 
\begin{gather} 
 \label{equ:ctcloss1} 
 \min_{\Theta_R} \mathcal{L}_{CTC}(\hat{t'}, t')    \\
 \label{equ:CEloss1} 
 \min_{\Theta_R, \Theta_T} (\mathcal{L}_{CTC}(\hat{t'}, t') +\mathcal{L}_{CE}(\hat{t}, t))
\end{gather}

\begin{figure}[t]
	\centering
    \includegraphics[width=0.85\columnwidth]{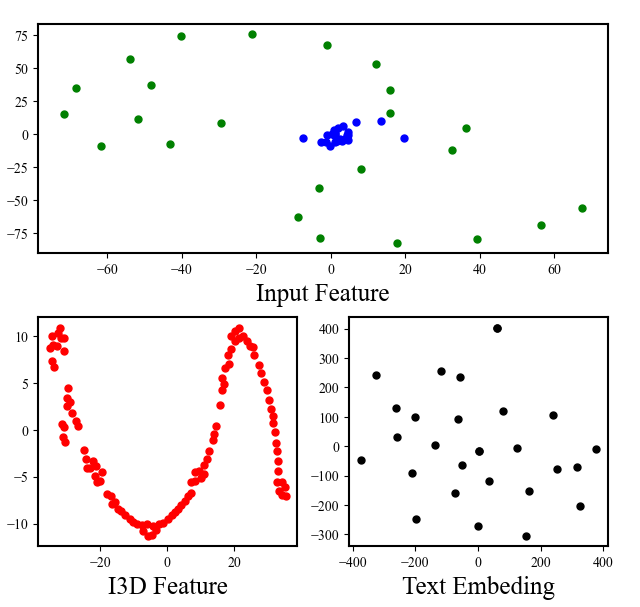}
 \caption{ Visualization of feature distribution of one test sample in How2Sign dataset.  Blue, green points represent the input features of translation model with/without TCP, and red and black points represent video features extracted by I3D model and corresponding  text sequence embedding.}
	\label{fig:tcp_how2sign} 
\end{figure} 

\textbf{The principle of TCP}:   
Typically, CTC loss requires that the source sequence and the target sequence are aligned in the same order. Thus, CTC loss is often used in CSLR task, where the sign video and gloss sequence have the same order. However, in SLT tasks, the alignment between the sign video and the spoken language sentence is usually non-monotonic, due to the difference of grammatical rules between sign language and spoken language. To tackle the above challenge of gloss-free SLT tasks, we design the Text-driven CTC Pre-training method, which generates pseudo gloss sequence from labeled sentence and then adopts CTC loss for model pre-training. 
Although the generated pseudo gloss sequences from text/sentences may be different from the ground-truth gloss sequences, our TCP is still crucial for improving SLT performance. The two possible reasons can be concluded as follows.
First,  given the discrete nature of languages, the inputs and outputs of translation models (\eg mBART used in SLT) are in a categorical, discrete-valued space (\ie typically as token embeddings~\cite{kudosentencepiece}). 
Second, without CTC pre-training, the features of adjacent frames/clips (extracted by the backbone) tend to remain in a continuous-valued space, while CTC pre-training with gloss sequences guides the segmentation of video features into discrete units.
Therefore, by using CTC with pseudo gloss sequence $t'$ to pre-train the backbone, TCP provides a weak tokenization function, even if it may result in inaccurate segmentation.

To verify the effectiveness of TCP, we visualize the feature distributions input to the translation model, \ie the features output from the CSLR model, while using or NOT using TCP. 
As shown in Figure~\ref{fig:tcp_vis}, the embedding distributions of gloss sequence (\ie red points) and words in spoken sentence (\ie black points) are normally in a discrete-valued space, as described in the first reason of the previous paragraph. 
When moving to the top part of Figure~\ref{fig:tcp_vis}, the features extracted by NOT using TCP locate at close positions (\eg blue points on a continuous curve), while the features extracted by using TCP scatter at different positions (\ie green points in a discrete-valued space), as described in the second reason of the previous paragraph. 
It demonstrates that introducing CTC loss with pseudo gloss annotations can guide the model to extract discrete and tokenized video features.

\textbf{The generalizability of TCP}:
It is worth mentioning that the TCP mechanism not only makes sense for our proposed model MixSignGraph, bus also works for other backbones. 
To verify this phenomenon, we also visualize the feature distribution of the other pre-trained backbone I3D, which is adopted for How2Sign and OpenASL datasets (see Section~\ref{sec:trainingsetting} for details). 
In Figure~\ref{fig:tcp_how2sign}, we show the output feature distribution of I3D backbone, the input feature distribution to the translation model by using or NOT using TCP, and the word embedding in spoken sentence. 
As shown in the bottom left part of Figure~\ref{fig:tcp_how2sign}, the features of I3D locate on a continuous curve (\ie features in adjacent time steps have high similarity). 
When moving to the top part of Figure~\ref{fig:tcp_how2sign}, we can find that the input features of the translation model by NOT using TCP (\ie blue points) gather together and are difficult to distinguish. 
However, when adopting TCP, the input features scatter in a discrete-valued space.  
From the perspective of feature distribution space, the pattern of feature distribution by using TCP is similar to that of word embedding in spoken sentence, \ie in a discrete-valued space.
The above phenomenon proves that TCP can also provide a weak tokenization function, even if TCP may bring inaccurate segmentation.

\section{Experiments}
\begin{table*}[t]
	\centering 
	\resizebox{0.95\textwidth}{!}{
\begin{tabular}{l|ccc|ccc|ccc|ccc} 
\multirow{2}*{Dataset}&\multicolumn{3}{c|}{Video Samples}& \multicolumn{3}{c|}{Gloss Vocabulary}&\multicolumn{3}{c|}{Token Vocabulary} &\multicolumn{3}{c}{Vocabulary in TCP} \\ 
\cline{2-13}
& {Train}&{Test}&{Validation}& {Train}&{Test}&{Validation}& {Train}&{Test}&{Validation} & {Train}&{Test}&{Validation}  \\ 
\toprule 
PHOENIX14~\cite{koller15:cslr}          &5672&629&540& 1103&497&462&-&-&-&-&-&-     \\
PHOENIX14T~\cite{cihan2018neural}        &7096&642&519& 1085&411&393& 2143&976&927&2888&1002&952\\
CSL-Daily~\cite{zhou2021improving} &18401&1176&1077&  2000&1345&1358  &2342&1358&1344 &2332&1351&1351\\ 
How2Sign~\cite{duarte2021how2sign}  &30904&2328&1713& -&-&-&8816&3312& 3046&9098&2573& 2302 \\ 
OpenASL~\cite{shi2022open}&96476&975&966&-&-&-& 13902&3336&3187&19144&2362&2220  
	\end{tabular}} 
	\caption{Details of datasets used in our paper.}  
	\label{tab:dataset}
\end{table*}

\subsection{Datasets} 
Our experiment evaluation is conducted on five publicly available SL datasets, including three widely used datasets (\ie PHOENIX, PHOENIX14T and CSL-Daily) and two new and large datasets (\ie How2Sign and OpenASL), as described below.
\begin{itemize}
    \item \textbf{PHOENIX14} \cite{koller15:cslr}: A widely used German SL dataset with 1295 glosses from 9 signers for CSLR. It includes 5672, 540 and 629 weather forecast samples for training, validation, and testing, respectively.
    
    \item \textbf{PHOENIX14T} \cite{cihan2018neural}:  A German SL dataset with both gloss annotations and translation annotations.  It contains 7096, 519 and 642 samples from 9 signers for training, validation and testing, respectively. In regard to the two-stage annotations, the sign gloss annotations have a vocabulary of 1066 different signs for CSLR, while the German translation annotations have a vocabulary of 2877 different words for SLT.
    
    \item \textbf{CSL-Daily} \cite{zhou2021improving}: A Chinese SL dataset with 18,401 labeled videos from 10 signers for training. It includes 2000 gloss annotations for CSLR and 2343 Chinese words for SLT.
    
    \item \textbf{How2Sign} \cite{duarte2021how2sign}:  A multimodal and multiview continuous American Sign Language (ASL) dataset. It consists of a parallel corpus, which has more than 80 hours of sign language videos and a set of corresponding modalities (\ie speech, English transcripts, and depth). 
    In the dataset, there are 31128, 1741 and 2322 samples for training, validation and testing, respectively. In regard to the annotations, there are no gloss-level annotations, thus the dataset is used for SLT tasks. 
    It is worth mentioning that our experiment only use the RGB modality (\ie NOT all modalities) with frontal view (\ie NOT all views) to train and test our model.
    
    \item \textbf{OpenASL} \cite{shi2022open}: A large ASL dataset with translation annotations for SLT. It includes 280 hours of ASL videos from more than 200 signers, and contains 96476, 997 and 999 samples for training, validation and testing, respectively.
\end{itemize} 

We provide detailed information of the five datasets in Table~\ref{tab:dataset}. In addition, we also provide the token vocabulary obtained by the mBART tokenizer and the vocabulary of pseudo gloss labels in TCP for reference. It is worth noting that the datasets we downloaded (\ie shown in Table~\ref{tab:dataset}) may slightly differ from the official versions described in their own papers. 

 \begin{table}[t]
\tablestyle{1.2pt}{1.2} 
	\centering
\resizebox{0.95\columnwidth}{!}{ 
		\begin{tabular}{cccccc|cc|cc}
\multirow{2}*{${LSG}_1$} &\multirow{2}*{${TSG}_1$}&\multirow{2}*{${HSG}_1$} &\multirow{2}*{${LSG}_2$} &\multirow{2}*{${TSG}_2$} &\multirow{2}*{${HSG}_2$} & \multicolumn{2}{c|}{Dev}& \multicolumn{2}{c}{Test} \\
&&&&&&WER&Del/Ins&WER&Del/ins\\
			\shline
\XSolidBrush & \XSolidBrush& \XSolidBrush&\XSolidBrush &\XSolidBrush &\XSolidBrush & 22.3 &8.4/2.5& 22.2 &8.1/2.7 \\ 
\Checkmark & & & & &  & 19.2&5.6/2.2&21.0&4.8/2.3 \\
&\Checkmark & &  & &   &19.6 &5.6/2.1 &21.5&5.1/2.5\\ 
& &\Checkmark &&&    &19.9 & 6.7/2.1&21.6&6.2/3.1\\
& & &\Checkmark&&   &19.3 &5.3/2.1 &20.8&5.8/2.0 \\            
& & && \Checkmark&  &19.5 &6.6/1.7  &21.2&5.4/2.4 \\
& & && &\Checkmark   &19.4&6.2/2.9&21.5&5.9/3.6\\
   \hline 
\Checkmark & & &\Checkmark& & &18.7 &5.1/2.3&20.6&5.2/1.7  \\
& \Checkmark & &  &  \Checkmark& &18.6 &4.3/1.8&20.2&5.5/1.7 \\
& &\Checkmark &  & &   \Checkmark &17.8 &6.1/2.3&19.6&5.3/2.9\\	
\Checkmark  &  \Checkmark& \Checkmark & & & &17.1 &5.3/2.0&19.7&5.0/3.0 \\
& & & \Checkmark  &  \Checkmark& \Checkmark &17.4 &5.7/2.1&19.6&5.1/2.9 \\	
    \hline
\Checkmark &\Checkmark & \Checkmark & \Checkmark& \Checkmark& \Checkmark & \textbf{16.8} &{5.1}/{2.0}&\textbf{19.6}& {5.0/3.2} \\	
	\end{tabular}}
	\caption{Effects of proposed sign graph modules.}
	\label{tab:signGraph}
\end{table}

\begin{table}[t]
 \tablestyle{3pt}{1.15}
	\centering
	\resizebox{0.95\columnwidth}{!}{ 
		\begin{tabular}{lll|cc|cc}
\multirow{2}*{${LSG}_1$} &\multirow{2}*{${TSG}_1$}&\multirow{2}*{${BSG}_1$}&  \multicolumn{2}{c|}{Dev}& \multicolumn{2}{c}{Test} \\
  & &&WER&Del/Ins&WER&Del/ins\\
			\shline
Dense &Sparse &Fixed  &\textbf{17.1} &{5.3/2.0} &\textbf{19.7}& {5
0/3.0} \\ 
Dense &Dense &Fixed   &19.3 &4.9/2.9 &20.1& 5.2/2.6\\
Sparse &Sparse &Fixed   &19.0 &5.3/2.1 &20.2&5.4/2.1\\
Sparse &Dense  &Fixed  &20.2 &5.3/2.6 &20.6&5.5/2.9\\ 
Dense &Sparse &Dynamic  &{18.1} &{5.1/2.2} &{20.1}& {5.1/2.1}\\ 
Dense &Dense &Dynamic   &19.9 &5.6/3.5 &21.8& 5.5/2.6\\
Sparse &Sparse &Dynamic   &19.1 &5.0/1.9 &21.1&5.8/2.1\\
Sparse &Dense  &Dynamic  &20.7 &6.0/2.1 &21.5&6.0/2.5\\ 
	\end{tabular}} 
	\caption{Effect of graph types.}
	\label{tab:GraphBuilding} 
\end{table} 

\subsection{Experimental Setting}
Here, we describe the baseline settings for our architecture.

\subsubsection{Data Preprocessing}
For data preprocessing, we adhere to the methodologies presented in the previous conference paper~\cite{gan2024signgraph}. During training, we apply data augmentations, including resizing frames to 256$\times$256 pixels, randomly cropping frames to 224$\times$224 pixels, random horizontal flipping with a probability of 0.5, and random temporal scaling (±20\%). During testing, frames are resized to 256$\times$256 pixels and center-cropped to 224$\times$224 pixels.

\subsubsection{Architecture Setting}
Our architecture is implemented using PyTorch 1.11. The setup includes the following components:
(1) \textit{Patchify stem}: Instead of using non-overlapping patches obtained via linear projection~\cite{dosovitskiy2020image}, which lacks the ability to model 2D local spatial context~\cite{wu2021cvt,xiao2021early}, we use a convolutional-based patchify stem (the first four stages of ResNet18) to obtain initial patch (node) embeddings. The dimension of the initial patches $D$ is set to 512.
(2) \textit{Patch merging}: Similar to the patchify stem, we use two convolutional blocks with a kernel size of $3\times3$ to downsample frames and obtain different-size patches after the $LSG$, $TSG$ and $HSG$ modules.
(3) \textit{Sign graph learning}: Our baseline includes two $LSG$, $TSG$ and $HSG$ modules. The initial values of $\mathsf{K}_{l}^1$ and $\mathsf{K}_{l}^2$ in the $LSG$ modules are set to 4, while that of $\mathsf{K}_{t}^1$ and $\mathsf{K}_{t}^2$ in the $TSG$ modules are set to 49 (corresponding to the number of nodes in $TSG_2$). 
In addition, we utilize the EdgeConv~\cite{wang2019dynamic} as the initial graph convolutional layer (GCN layer).
(4) \textit{Distance function}: In the baseline setting, we measure the distance between two nodes by using the Euclidean distance.
(5) \textit{Global feature module}: This module comprises two 1D convolution blocks, a 2-layer BiLSTM with a hidden size of 1024 for global feature modeling, and a fully connected layer for the final prediction.
(6) \textit{Translation network}: For a fair comparison, following the current SOTA SLT model~\cite{chen2022two}, we adopt the pre-trained mBART model provided by huggingface\footnote{https://huggingface.co/facebook/mbart-large-cc25} as our translation network.

\subsubsection{Training Setting}
\label{sec:trainingsetting}
For fair comparisons, we adopt the same training settings as in previous work~\cite{min2021visual}. The model is trained using the Adam optimizer with a weight decay of 0.0001 for 50 epochs on three GeForce RTX 3090 GPUs. The initial learning rate is set to 0.0001 for the recognition model and 0.00001 for the translation model. The learning rate is reduced by a factor of 0.5 at the 20th and the 30th epoches.
The batch size is set to 6. The CTC loss is applied after both 1D convolution blocks and the fully connected layer as the loss function. 
In regard to the large-scale datasets (\ie How2Sign and OpenASL), we adopt the visual features generated by a pre-trained I3D model~\cite{tarres2023sign} instead of fully training MixSignGraph to extract features, when considering the insufficient training memory and excessively long training time (\eg more than one month to train MixSignGraph on OpenASL).
In addition, to ensure the model can be trained end-to-end with the 24GB memory GPU, the entire model is trained in half-precision.


\begin{table}[t]
\tablestyle{2pt}{1.2} 
	\centering
	\resizebox{0.95\columnwidth}{!}{ 
		\begin{tabular}{c|cc|cc}
\multirow{2}*{Model} & \multicolumn{2}{c|}{Dev}& \multicolumn{2}{c}{Test} \\
&WER&Del/Ins&WER&Del/ins\\
			\shline
HSG $\rightarrow$ LSG $\rightarrow$ TSG & {16.8} &{5.1}/{2.0}&{19.6}& {5.0/3.2}\\  
HSG $\rightarrow$ TSG $\rightarrow$ LSG & \textbf{16.7} &4.9/2.1   & \textbf{19.0} &4.9/2.9 \\ 
LSG $\rightarrow$ HSG $\rightarrow$ TSG & 17.2 &6.3/2.0 & 19.8 & 4.8/2.7\\ 
LSG $\rightarrow$ TSG $\rightarrow$ HSG & 17.1 &5.4/2.0  & \textbf{19.0} &5.0/2.7 \\ 
TSG $\rightarrow$ HSG $\rightarrow$ LSG & 16.9 &5.5/1.8   & 19.3 &5.0/3.2 \\ 
TSG $\rightarrow$ LSG $\rightarrow$ HSG & 16.9 &5.1/2.1   & 19.5 &4.9/3.2 \\ 
	\end{tabular}} 
	\caption{Effect of order of proposed sign graph modules.}
	\label{tab:GraphOrder}
\end{table}

\begin{figure}[t]
	\centering
\includegraphics[width=0.48\columnwidth]{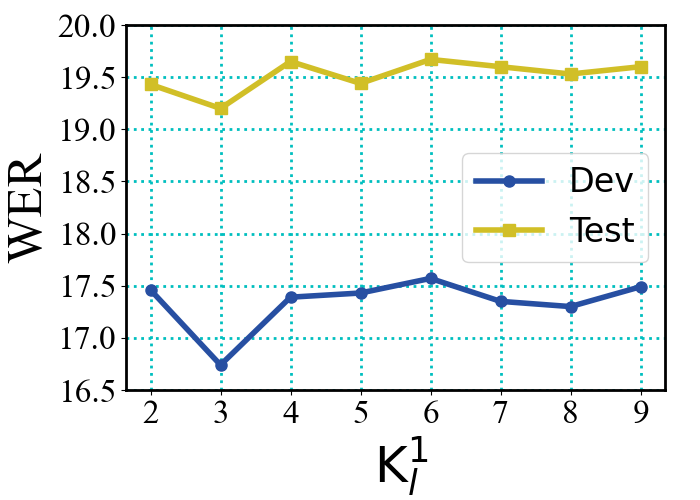}	\includegraphics[width=0.48\columnwidth]{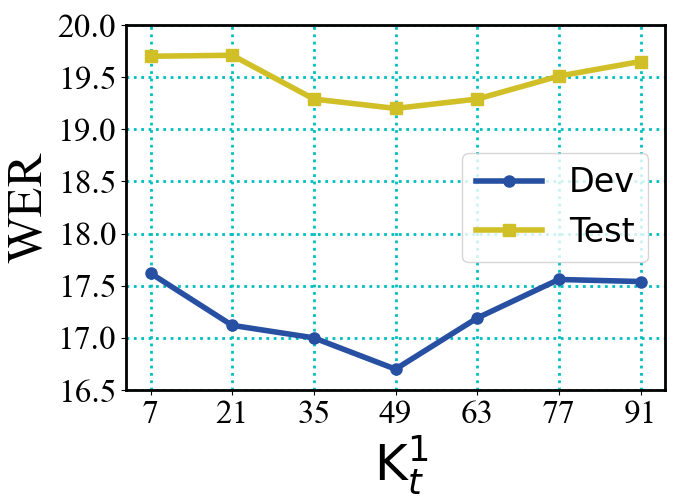}
\includegraphics[width=0.48\columnwidth]{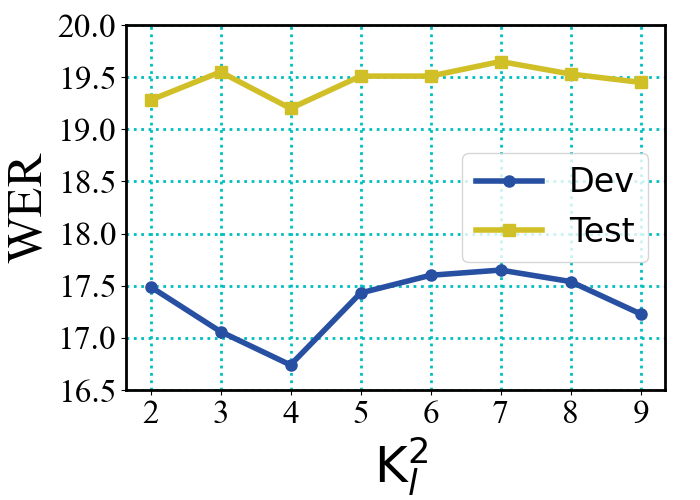}
\includegraphics[width=0.48\columnwidth]{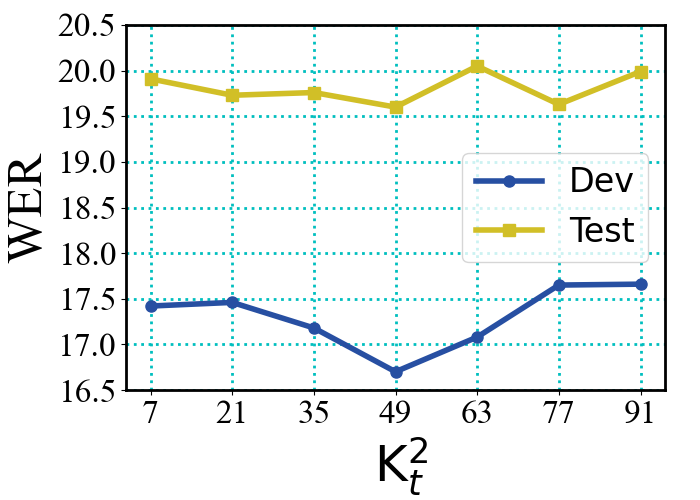}
 \caption{Effects of $\mathsf{K}_l$ and $\mathsf{K}_t$.}
	\label{fig:k1}
\end{figure}

\subsubsection{Evaluation Tasks}  
We evaluate the performances of our model on the following tasks:
\label{Sec:EvaluTask}
\begin{itemize} 
\item \textbf{CSLR:} Recognizing a sign sequence as a corresponding gloss sequence, also known as Sign2Gloss.

\item \textbf{SLT}: Translating a sign sequence into a spoken sentence, and it can be classified into the following three categories:
\begin{itemize}

\item \textbf{Sign2Gloss2Text:} First, the recognized gloss sequence is obtained based on the CSLR model. Then, the predicted glosses are translated to a spoken sentence by a translation model. 

\item \textbf{Sign2Text:} Directly translating a sign language video into a spoken sentence. Gloss annotations are required for pre-training a CSLR model, and visual features output by the CSLR model are send to a translation model for generating the spoken sentence.

\item \textbf{Gloss-free Sign2Text:} Directly translating a sign language video into a spoken sentence, while NOT using gloss annotations to pertrain a CSLR model. This task is also known as gloss-free SLT. 
\end{itemize}
\end{itemize}

\subsubsection{Evaluation Metrics} To evaluate our model, we adopt the Word Error Rate (WER) metric for CSLR task, while adopting the ROUGE-L F1 Score~\cite{lin2004rouge} and BLEU-1,2,3,4~\cite{papineni2002bleu} for all SLT tasks, including Sign2Gloss2Text, Sign2Text, and gloss-free SLT. These metrics have often been used to measure the quality of recognition~\cite{gan23contrastive, min2021visual} and translation in existing work~\cite{gan2021skeleton, chen2020simple, chen2022two}.

\begin{table*}[t]
    \vspace{-1em}
    \centering
    \subfloat[
    \textbf{Backbone}. Comparison of different backbones.
    \label{tab:ablation:backbone} ]{
        \centering
        \begin{minipage}{0.29\linewidth}{\begin{center}
        \tablestyle{2pt}{1.1}
        \begin{tabular}{z{32}x{24}x{24}x{24}}
         BackBone& WER & Del & Ins   \\
        \shline
        SwinT~\cite{liu2021swin}& 45.4&   16.3& 1.3   \\
        PyVIG~\cite{han2022vision} &35.4&  12.1 &  1.3 \\ 
        SA~\cite{vaswani2017attention}   &39.2&  15.3&  0.9   \\ 
        Ours &  \textbf{16.7} &{5.1} &{2.0} \\
        \end{tabular}
        \end{center}}\end{minipage}
    }
\hspace{2em}
    \subfloat[ \textbf{Distance function}. Comparison of distance functions. \label{tab:ablation:distance} ]{
        \begin{minipage}{0.29\linewidth}{\begin{center}
        \tablestyle{4pt}{1.1}
        \begin{tabular}{z{36}x{24}x{24}x{24}}
        Distance & WER & Del & Ins\\
        \shline
        Cosine & 17.4	&5.5 & 2.0 \\ 
        Chebyshev  &17.3&5.1&3.2 \\ 
        Euclidean & \textbf{16.7} &{4.9} &{2.1}\\
        \multicolumn{2}{c}{} \\
        \end{tabular}
        \end{center}}\end{minipage}
    }
\hspace{2em}
\subfloat[\textbf{Graph Convolution} Effect of different GCN layers. \label{tab:ablation:graphconv}]{
    \begin{minipage}{0.29\linewidth}{\begin{center}
        \tablestyle{2pt}{1.1}
        \begin{tabular}{z{52}x{36}x{24}x{24}}
            \multicolumn{1}{c}{GraphConv} & WER& Del & Ins \\
            \shline GATv2Conv~\cite{brody2021attentive} & 17.0 & 5.1& 2.0\\
        SAGEConv~\cite{hamilton2017inductive} &17.7 &5.9 &1.9 \\
            GCNConv~\cite{kipf2016semi} & {17.7} &{5.5} &{2.0}\\ 
             EdgeConv~\cite{wang2019dynamic} &\textbf{16.7} &{4.9} &{2.1}\\
            \end{tabular} 
        \end{center}}
    \end{minipage}
}
\\
\centering 
\subfloat[ \textbf{Patch size}. Comparison of different patch sizes with one $LSG$, $TSG$ and $BSG$ modules.\label{tab:ablation:patchszie} ]{
    \begin{minipage}{0.29\linewidth}{
    \begin{center}
    \tablestyle{2pt}{1.1}
    \begin{tabular}{x{30}x{24}x{24}x{24}}
     PatchSize& WER & Del & Ins  \\
    \shline
    8 &17.4  &5.4&3.0\\
    16 & 17.1&5.3&2.0\\
    32  &17.5&5.7&2.4\\
    \end{tabular}
    \end{center}}
    \end{minipage}
}
\hspace{2em}
\subfloat[ \textbf{Multiscale SignGraph}. Effect of the number of stages in multiscale SignGraph. \label{tab:ablation:mutilscale}]{
	\centering
	\begin{minipage}{0.29\linewidth}{\begin{center}
	\tablestyle{4pt}{1.1}
	\begin{tabular}{x{40}x{24}x{20}x{20}}
	Stages& WER & Del &Del\\
	\shline 
	16$\rightarrow$32 &  \textbf{16.7} & {4.9} & {2.1} \\
 	8$\rightarrow$16$\rightarrow$32 & 17.1 &5.2 &2.0 \\ %
  4$\rightarrow$8$\rightarrow$16$\rightarrow$32&  17.8&5.4&3.1\\
	\end{tabular}
	\end{center}}\end{minipage}
}
\hspace{2em}
\subfloat[ \textbf{Drop edge}. Adding DropEdge~\cite{rong2020dropedge} in the local graph and hierarchical graph does not improve performance. \label{tab:ablation:dropedge} ]{
	\begin{minipage}{0.29\linewidth}{
        \begin{center}
	\tablestyle{4pt}{1.1}
    	\begin{tabular}{x{48}x{24}x{24}x{24}}
    	DropRate& WER & Del & Ins \\
    	\shline
    	0 & \textbf{16.7} & {4.9} & {2.1} \\
    	15\% & 17.7 & 6.1 &1.4\\
            30\% & 17.5 & 6.0 &2.1\\   
	\end{tabular}
	\end{center}}\end{minipage}
}  
\caption{Ablation experiments of SignGraph on PHOENIX14T dev set.}
\label{tab:ablations} 
\end{table*}

\begin{table*}[t]
\tablestyle{2pt}{1.2} 
	\centering
	\resizebox{0.97\textwidth}{!}{ 
		\begin{tabular}{c|c|ccccc|ccccc|cc|cc}
\multirow{2}*{Dataset}&\multirow{2}*{Model} & \multicolumn{5}{c|}{Dev}& \multicolumn{5}{c|}{Test} &\multicolumn{2}{c|}{Dev}& \multicolumn{2}{c}{Test}  \\
&&ROUGE&BLEU1&BLEU2&BLEU3&BLEU4&ROUGE&BLEU1&BLEU2&BLEU3&BLEU4 &WER&Del/Ins&WER&Del/Ins\\
			\shline 
\multirow{2}*{PHOENIX14T}
&w/o TCP &32.63 &33.57&18.71& 13.47 & 9.20 &34.56 & 35.31& 21.05&16.45&9.40&-&-/-&-&-/- \\ 
&w/ TCP& \baseline{51.71} &\baseline{51.07}&\baseline{37.97}  &\baseline{29.98}&\baseline{24.87}&\baseline{51.14} &\baseline{50.01}&\baseline{38.04}  &\baseline{29.95}&\baseline{24.02}&59.97&36.8/2.5&59.55&35.9/2.7\\  
&w/ gloss&\better{55.77}&\better{55.01}&\better{42.64}& \better{34.94}&\better{29.00}&\better{53.84}&\better{54.90}&\better{42.53}&\better{34.50}&\better{28.97}&{16.72} &4.9/2.1   & {19.01} &4.9/2.9\\
\hline
\multirow{2}*{CSL-Daily}
&w/o TCP & 34.11& 33.78  &20.13&12.61 &8.01&32.21&32.86&18.34&10.67&6.78 \\ 
&w/ TCP& \better{49.16}&\better{49.98}&\better{36.42}&\better{26.89}&\better{20.43}&\better{49.93}&\better{50.24}&\better{36.91}&\better{27.54}&\better{21.01}&66.87&39.11/4.9&66.05&38.18/4.9  \\  
&w/ gloss &\better{54.54}&\better{55.87}&\better{42.45}&\better{32.75}&\better{25.77}&\better{54.67}&\better{ 55.41}&\better{42.43}&\better{32.84}&\better{25.87}&{25.13}&{6.4/2.1}&{25.01}&{7.0/1.6}\\
\hline
\multirow{2}*{How2Sign}
&w/o TCP &18.35 &23.88   &12.92 &7.89 & 5.05&18.06&23.41&12.55&7.61&4.86&-&-/-&-&-/-  \\ 
&w/ TCP& \better{29.24} &\better{34.82} &\better{22.47} &\better{15.61} &\better{11.28}&\better{28.01}&\better{32.74}&\better{20.83}&\better{14.41}&\better{10.41}&72.38&46.35/2.69&74.88&50.27/2.07 \\   
\hline
\multirow{2}*{OpenASL}
&w/o TCP &12.86 &11.95 &4.89 &2.78 &1.88&12.56 &11.29&4.64&2.77&1.95&-&-/-&-&-/- \\ 
&w/ TCP& \better{25.41}& \better{26.82}& \better{16.70}& \better{11.48}& \better{8.36}& \better{25.71} &\better{26.65}&\better{16.55}&\better{11.68}&\better{8.69}&{81.53}& {59.88/1.08} & {81.33}&{60.39/1.29}\\  
	\end{tabular}} 
	\caption{Effect of proposed TCP for SLT. Besides, we also show the CSLR performance based on pseudo gloss sequence obtained by text labels in the right part.}
	\label{tab:TCP_other}
\end{table*}

\subsection{Ablation Study}\label{sec:ablation}
Following the previous work~\cite{cihan2018neural}, we perform ablation studies on PHOENIX14T dataset to verify the effectiveness of the proposed MixSignGraph model. 
 

\subsubsection{Effects of Proposed Sign Graph Modules} To demonstrate the effectiveness of the three proposed graph modules (\ie LSG, TSG, HSG), we show the CSLR performance of MixSignGraph while removing some of these graph modules, as shown in Table~\ref{tab:signGraph}.
Results indicate that all the proposed $LSG$, $TSG$, $HSG$ modules contribute to a higher performance. 
Specifically, our baseline which only utilizes ResNet18 as the backbone achieves 22.3\% WER on PHOENIX14T dev set.
When only adopting $LSG_1$, $LSG_2$, ${TSG}_1$, ${TSG}_2$,  ${HSG}_1$ or ${HSG}_2$ module, WER on dev set decreases by 3.1\%, 3.0\%,  2.7\%, 2.8\% 2.4\% or 2.9\%, respectively.
The results prove that all the $LSG$, $TSG$ and $HSG$ modules can effectively learn sign-related features, and the learned intra-frame cross-region features, inter-frame cross-region features and different-granularity one-region features are crucial for a higher CSLR performance.
\begin{figure*}[t]
	\centering
\includegraphics[width=0.7\textwidth]{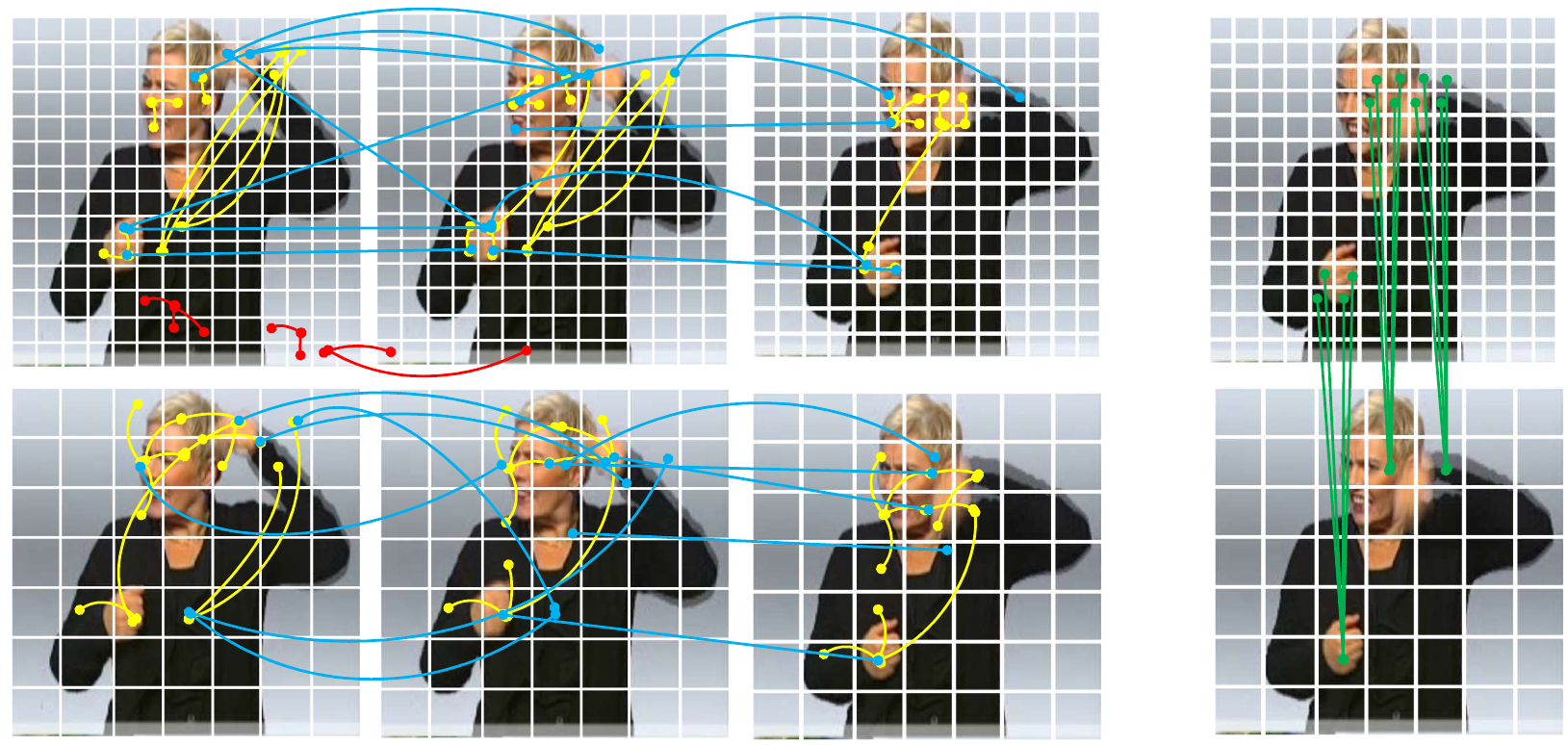} 
	\caption{Visualization of graph construction: $LSG$ and $TSG$ modules ($LSG_1$ and $TSG_1$ in the first row, $LSG_2$ and $TSG_2$ in the second row) are in the left part, $HSG$ module is in the right part. The graph in $LSG$, $TSG$ and $HSG$ module is shown in yellow, blue and green, respectively. We also show some ‘unimportant’ edges between nodes of the background in red color.} 
	\label{fig:graphbuilding}
 \vspace{-4mm}
\end{figure*}
In addition, when we combine the designed $LSG_1$ and $LSG_2$ modules, ${TSG}_1$ and ${TSG}_2$ modules, ${HSG}_1$ and ${HSG}_2$ modules, WER on dev set decreases by 3.6\%, 3.7\%, 4.5\%, respectively. 
When we replace the backbone with designed ${LSG}_1$, ${TSG}_1$, ${HSG}_1$ modules, or ${LSG}_2$, ${TSG}_2$, ${HSG}_2$ modules, WER decreases by 5.2\% or 4.9\%, respectively.  
Besides, when we combine all proposed modules, the CSLR performance can achieve further improvement, \ie WER decreases by 5.5\%/2.6\% on dev/test set.
It indicates that the combination of mixed graph modules can further boost the CSLR performance.

\subsubsection{Effect of Graph Type} In MixSignGraph, we build dynamic dense graphs (\ie connecting the top $\mathsf{K}$ neighbors for each node) in the $LSG$ module, dynamic sparse graphs (\ie connecting only the top $\mathsf{K}$ node pairs) in the $TSG$ module, and fixed graphs (\ie connecting the same regions between feature maps) in the $HSG$ module. 
To verify the effectiveness of the above graph types, we evaluate the CSLR performance by changing the graph types in $LSG$, $TSG$ and $HSG$ modules. 
For simplicity, we only use one $LSG$ module, one $TSG$ module and one $HSG$ module in the experiment.
As shown in Table~\ref{tab:GraphBuilding}, when using a spare graph in the $LSG$ module, a dense graph in the $TSG$ module, or a dynamic dense graph (\ie building in the same way with the $LSG$ module) in the $HSG$ module, there is a noticeable performance drop. That is to say, it is effective to adopt dense, sparse and fixed graphs in the $LSG$, $TSG$ and $HSG$ module, and the combination of these mixed graphs can make our model achieve the best recognition performance. 

\subsubsection{Effect of Ordering of Proposed Sign Graph Modules} 
We also evaluate the CSLR performance, by changing the ordering of three graph modules. As shown in Table~\ref{tab:GraphOrder}, different orderings of the proposed graph modules lead to slight differences in the CSLR performance, \ie the ordering has a little effect on the overall effectiveness of feature extraction and integration. Nevertheless, the ordering $HSG$ $\rightarrow$ $TSG$ $\rightarrow$ $LSG$ can help the model to achieve the best CSLR performance. Thus our model MixSignGraph adopts the ordering $HSG$ $\rightarrow$ $TSG$ $\rightarrow$ $LSG$ by default.

\subsubsection{Effect of Hyperparameter $\mathsf{K}$} 
In the proposed sign graph modules, there are four hyperparameters: $\mathsf{K}_l^1$ and $\mathsf{K}_l^2$ in the $LSG$ modules, $\mathsf{K}_t^1$ and $\mathsf{K}_t^2$ in the $TSG$ modules. 
To determine the best values for these hyperparameters, we evaluate the CSLR performance by changing $\mathsf{K}_l$ (\ie both $\mathsf{K}_l^1$ and $\mathsf{K}_l^2$), $\mathsf{K}_t$ (\ie both $\mathsf{K}_t^1$ and $\mathsf{K}_t^2$) values from 2 to 9, 7 to 91, respectively. 
However, due to the large number of possible combinations of $\mathsf{K}_l^1$, $\mathsf{K}_l^2$, $\mathsf{K}_t^1$ and $\mathsf{K}_t^2$, it is impractical to make an exhaustive search to find the globally optimal values of these hyperparameters. 
Therefore, we fix three of the parameters at a time and adjust the fourth to find suitable values. 
Finally, based on the WER performance on the DEV set, we set $\mathsf{K}_l^1$, $\mathsf{K}_l^2$, $\mathsf{K}_t^1$ and $\mathsf{K}_t^2$ to 3, 4, 49 and 49, respectively, as shown in Figure~\ref{fig:k1}. 

\subsubsection{Effects of Different Backbones} 
To verify the effectiveness of the proposed MixSignGraph model, we replace our backbone with other patch-based networks (\eg CvT~\cite{xiao2021early}, Swin Transformer~\cite{liu2021swin}), or replace our graph modules with Self-Attention (SA) layers~\cite{vaswani2017attention}.   
As shown in Table~\ref{tab:ablation:backbone}, simply adopting the patch-based SOTA backbones or substituting our graph modules with SA layers does not lead to appealing performances on the CSLR task.
In contrast, our proposed MixSignGraph backbone achieves a satisfactory performance, by effectively capturing both cross-region and one-region features. 

\subsubsection{Effect of Distance Function}  
In the MixSignGraph, we use the KNN algorithm to find the nearest neighbors for a node/patch based on the distance between two nodes. To select a suitable distance function to measure the distance between nodes, we evaluate the CSLR performance with the following commonly used distance functions, \ie Cosine distance, Chebyshev distance and Euclidean distance. As shown in Table~\ref{tab:ablation:distance}, there are subtle performance differences, when using different distance functions. In MixSignGraph, we adopt Euclidean distance for a better performance. 

\begin{CJK}{UTF8}{gbsn} 

\begin{table*}[t]
	\centering 
	\begin{minipage}[b]{0.9\textwidth}
		\centering
	\end{minipage}
	\resizebox{0.9\textwidth}{!}{
		\centering
		\begin{tabular}{l|l}
			\toprule 
              example(a)&PHOENIX14 dataset\\
        \hline
        Groundtruth& MORGEN DASSELBE SCHAUER REGION SONST VIEL SONNE REGION TEILWEISE WEHEN STARK\\
        MultiSignGraph &  MORGEN {\color{blue}DASSELBE}  SCHAUER REGION SONST VIEL SONNE REGION TEILWEISE WEHEN {\color{green}SCHWACH} \\
        MixSignGraph & MORGEN DASSELBE SCHAUER REGION SONST VIEL SONNE REGION TEILWEISE WEHEN STARK \\  
        \hline
           example(b)&PHOENIX14T dataset\\
        \hline
        Groundtruth&     DARUNTER NEBEL LANG IN-KOMMEND DANEBEN SONNE BERG OBEN DANN DURCHGEHEND SONNE\\
        MultiSignGraph & {\color{red}\_\_ON\_\_} DARUNTER NEBEL LANG IN-KOMMEND DANEBEN SONNE {\color{blue}BERG} OBEN DANN DURCHGEHEND SONNE\\
        MixSignGraph &  DARUNTER NEBEL LANG IN-KOMMEND DANEBEN SONNE {\color{blue}BERG} OBEN DANN DURCHGEHEND SONNE\\ 
        \hline
        example(b)&CSL-Daily dataset\\
        \hline
        Groundtruth& 他 分别 年 十 学生 见面 （he separate year ten student meeting）\\
        MultiSignGraph & 他 分别 {\color{blue}年} 十 同学 {\color{red}没有} 见面\\
        MixSignGraph &  他 分别 {\color{blue}年} 十 同学 见面 \\  
         
			\bottomrule	 	
	\end{tabular}}
  \captionof{table}{CSLR Qualitative results on PHOENIX14, PHOENIX14T and CSL-Daily.
We use different colors to represent {\color{green}substitutions}, {\color{blue}deletions}, and {\color{red}insertions}, respectively.}
	\label{tab:SLRQualitative} 
\end{table*}

\begin{table*}[t]
	\centering 
	\begin{minipage}[b]{0.9\textwidth}
		\centering
	\end{minipage}
	\resizebox{0.9\textwidth}{!}{
		\centering
		\begin{tabular}{l|l}
			\toprule 
           example(a)&PHOENIX14T dataset\\
        \hline
        Groundtruth&und zum wochenende wird es dann sogar wieder ein bisschen kälter\\
        w/o TCP &und zum wochenende  teil recht freundliche hochdruckwetter begleitet uns\\
        w/ TCP & und zum wochenende wird es dann sogar ein bisschen kälter\\ 
        Gloss-based & und zum wochenende wird es dann sogar wieder ein bisschen kälter\\
        \hline
        example(b)&CSL-daily dataset\\
        \hline
        Groundtruth&这 件 事 分 别 有 什 么 好 处 和 坏 处 ？\\
        w/o TCP &这 事 有 什 么 好 处 和 坏  处？\\
        w/ TCP & 这 事 {\color{red}情} 分 别 有 什 么 好 处 和 坏 处？\\ 
        Gloss-based & 这 件 事 分 别 有 什 么 好 处 和 坏 处 ？\\
        \hline
	example(c)&How2Sign dataset\\
        \hline
        Groundtruth & You can take this forward and back, you can take it in a circle, you can take it in a lot of different directions.\\
        w/o TCP & Take a step forward, back; you can do a circular mothion that can go from on direction to the other.\\
        w/ TCP &You can take this forward and back, you can take it in {\color{red}circles}, you can take it in a lot of different {\color{red}ways}.\\

        \hline
        example(d)&OpenASL dataset\\
        \hline
        Groundtruth& There are results pending for 20 other tests\\ 
        w/o TCP & There are {\color{red}now fires and reports} other {\color{red}people injured}\\
        w/ TCP &   There are {\color{red}waiting} for the results of {\color{red}the} 20 other tests\\
			\bottomrule	 	
	\end{tabular}}
  \captionof{table}{SLT  Qualitative results on PHOENIX14T, CSL-Daily, How2Sign and OpenASL.}
\label{tab:SLTQualitative} 
\end{table*} 
\end{CJK}

\subsubsection{Effect of GCN Layer} 
To verify the effectiveness of the graph convolution layer, we test the CSLR performance by adopting the following representative variants of graph convolution, \ie GATv2Conv~\cite{brody2021attentive}, SAGEConv~\cite{hamilton2017inductive}, GCNConv~\cite{kipf2016semi}, and EdgeConv~\cite{wang2019dynamic}. 
As shown in Table~\ref{tab:ablation:graphconv}, even using different GCN layers, our model still achieves good performance. It indicates that our MixSignGraph model has a good flexibility in GCN layer selection. Nevertheless, among the above GCN layers, EdgeConv can help our model to achieve the best CSLR performance, \ie 16.7\% WER. Therefore, we adopt the EdgeConv layer in MixSignGraph for a better performance. 

\subsubsection{Effect of Patch Sizes} 
Patches with different sizes have different receptive fields, thus can capture different-range features and further affect the model performance. 
To observe the effect of patch sizes, we evaluate the CSLR performance by only adopting one $LSG$ module, one $TSG$ module and one $HSG$ module, while changing the patch size. 
As shown in Table~\ref{tab:ablation:patchszie}, when using a patch size that is too small (\ie 8$\times$8) or too large (\ie 32$\times$32), the model MixSignGraph shows a worse performance. We found that an suitable patch size strikes the best balance, leading to optimal performance. 

\subsubsection{Effect of Multiscale Sign Graphs} 
To leverage the scale-invariant property of images~\cite{han2022vision}, PyVIG adopts a pyramid architecture that gradually increases patch size from 4 to 32 by shrinking the spatial size of feature maps. 
Similarly, to extract better sign features and achieve higher CSLR performance, we gradually add $LSG$, $TSG$ and $HSG$ modules after the early stage of the patchify stem.
According to Table~\ref{tab:ablation:mutilscale}, adding more $LSG$, $TSG$ and $HSG$ modules does not always bring performance gains. 
Therefore, MixSignGraph adopts two $LSG$, $TSG$ and $HSG$ modules, \ie increasing patch size from 16 to 32, aiming to achieve the best performance. 

\subsubsection{Effect of DropEdge} 
DropEdge~\cite{rong2020dropedge} can alleviate the over-smoothing and overfitting problems in dense graphs, by randomly removing a certain number of edges from the input graph at each training epoch.
Considering that we build dense graphs for the $LSG$ and $HSG$ modules, we introduce DropEdge into these two modules and change the drop rates of DropEdge to evaluate the CSLR performance. 
As shown in Table~\ref{tab:ablation:dropedge}, adding DropEdge does not improve the performance of our MixSignGraph. It indicates that our model can achieve excellent performance without relying on external modules (\eg DropEdge) and it can handle over-smoothing and overfitting problems well. 

\subsubsection{Effect of TCP for gloss-free SLT} 
To demonstrate the effectiveness of our proposed TCP mechanism, we compare a SLT model trained end-to-end (\ie NOT using TCP) and the same model pre-trained with TCP, in terms of SLT performance.
As shown in Table~\ref{tab:TCP_other}, our model pre-trained with TCP significantly outperforms the one trained end-to-end, \eg our model improves gloss-free SLT performance by 19.58 ROUGE score and 15.67 BLEU4 score on PHOENIX14T dev set. 
In addition, on PHOENIX14T and CSL-Daily datasets, the performance of our model pre-trained with TCP is very close to that of the gloss-based SLT model (\ie pre-trained with gloss annotations), \eg 52.21 ROUGE score vs 55.77 ROUGE score on PHOENIX14T. It indicates that the proposed TCP mechanism can substantially bridge the performance gap between gloss-free SLT and gloss-based SLT. 
\begin{table*}[t]
\centering 
\resizebox{0.95\textwidth}{!}{
\begin{tabular}{l|c|cccc|cccc|cc}
\multirow{3}{*}{Model} & \multirow{3}{*}{Backbone} & \multicolumn{4}{c|}{Extra cues} & \multicolumn{4}{c|}{PHOENIX14} & \multicolumn{2}{c}{PHOENIX14T} \\
&& & & & & \multicolumn{2}{c}{DEV} & \multicolumn{2}{c|}{TEST} & DEV & TEST \\ 
\cline{3-12}
&& {F/M} & {H} & {S} & {P} & WER & $\text{del}/\text{ins}$ & WER & $\text{del}/\text{ins}$ & WER & WER \\
\shline
STMC~\cite{zhou2020spatial} & VGG11 & $\checkmark$ & $\checkmark$ & $\checkmark$ & & 21.1 & $\text{7.7}/\text{3.4}$ & 20.7 & $\text{7.4}/\text{2.6}$ & 19.6 & 21.0 \\
C2SLR~\cite{zuo2022c2slr} & ResNet18 & & & $\checkmark$ & & 20.5 & $\text{-}/\text{-}$ & 20.4 & $\text{-}/\text{-}$ & 20.2 & 20.4 \\
TwoStream~\cite{chen2022two} & S3D & $\checkmark$ & $\checkmark$ & $\checkmark$ & $\checkmark$ & 18.4 & $\text{-}/\text{-}$ & 18.8 & $\text{-}/\text{-}$ & 17.7 & 19.3 \\
CrossL-Two~\cite{wei2023improving} & S3D & $\checkmark$ & $\checkmark$ & $\checkmark$ & $\checkmark^{*}$ & \baseline{15.7} & $\text{-}/\text{-}$ & \baseline{16.7} & $\text{-}/\text{-}$ & \baseline{16.9} & \baseline{18.5} \\
CrossL-Single~\cite{wei2023improving} & S3D & & & & $\checkmark^{*}$ & {-} & $\text{-}/\text{-}$ & {-} & $\text{-}/\text{-}$ & 20.6 & 21.3 \\
RTG-Net~\cite{gan2023towards} & RepVGG & $\checkmark$ & $\checkmark$ & $\checkmark$ & $\checkmark$ & 20.0 & $\text{8.4}/\text{1.5}$ & 20.1 & $\text{8.6}/\text{1.7}$ & 19.63 & 20.01 \\
\hline
Joint-SLRT~\cite{camgoz2020sign} & GooleNet & & & & $\checkmark$ & - & - & - & - & 24.6 & 24.5 \\
TwoStream~\cite{chen2022two} & S3D & & & & $\checkmark^{*}$ & 22.4 & $\text{-}/\text{-}$ & 23.3 & $\text{-}/\text{-}$ & 21.1 & 22.4 \\
VAC~\cite{min2021visual} & ResNet18 & & & & $\checkmark$ & 21.2 & $\text{7.9}/\text{2.5}$ & 22.3 & $\text{8.4}/\text{2.6}$ & - & - \\
SMKD~\cite{hao2021self} & ResNet18 & & & & $\checkmark$ & 20.8 & $\text{6.8}/\text{2.5}$ & 21.0 & $\text{6.3}/\text{2.3}$ & 20.8 & 22.4 \\
CorrNet~\cite{hu2023continuous} & ResNet18 & & & & $\checkmark$ & 18.8 & $\text{5.6}/\text{2.8}$ & 19.4 & $\text{5.7}/\text{2.3}$ & 18.9 & 20.5 \\
FCN~\cite{cheng2020fully} & customed & & & & & 23.7 & $\text{-}/\text{-}$ & 23.9 & $\text{-}/\text{-}$ & - & - \\
Contrastive~\cite{gan23contrastive} & ResNet18 & & & & & 19.6 & $\text{5.1}/\text{2.7}$ & 19.8 & $\text{5.8}/\text{3.0}$ & 20.0 & 20.1 \\
\hline
HST-GNN~\cite{kan2022sign} & Customized(GCN) & $\checkmark$ & $\checkmark$ & $\checkmark$ & $\checkmark$ & 19.5 & $\text{-}/\text{-}$ & 19.8 & $\text{-}/\text{-}$ & 19.5 & 19.8 \\
CoSign~\cite{jiao2023cosign} & ST-GCN(GCN) & $\checkmark$ & $\checkmark$ & $\checkmark$ & & 19.7 & $\text{-}/\text{-}$ & 20.1 & $\text{-}/\text{-}$ & 19.5 & 20.1 \\
\baseline{MultiSignGraph} & Customized(GCN) & & & & $\checkmark$ & 18.2 & $\text{4.9}/\text{2.0}$ & 19.1 & $\text{5.3}/\text{1.9}$ & 17.8 & 19.1 \\
\baseline{MixSignGraph} & Customized(GCN) & & & & $\checkmark$ & \baseline{16.5} & $\text{4.9}/\text{2.0}$ & \baseline{17.3} & $\text{4.9}/\text{2.2}$ & \baseline{16.7} & \baseline{19.0} \\
\end{tabular}}  
\caption{Comparison of CSLR performance on PHOENIX14 and PHOENIX14T datasets. (F: face, M: mouth, H: hands, S: skeleton, P: pre-training backbone with ImageNet, $\checkmark^{*}$: pre-training on other sign language datasets. Same applies to the table below.)} 
\label{tab:Phoenix}
\end{table*}

\begin{table}[t]
\tablestyle{1pt}{1.1}
\centering 
\resizebox{0.95\columnwidth}{!}{
\begin{tabular}{*l|c|cc|cc|cc}
\multirow{2}{*}{CSLR} & \multirow{2}{*}{Backbone} & \multicolumn{2}{c|}{Extra cues} & \multicolumn{2}{c|}{DEV} & \multicolumn{2}{c}{TEST} \\ 
& & S & P & WER & del/ins & WER & del/ins \\ 
\shline
Joint-SLRT~\cite{camgoz2020sign} & GoogleNet & & $\checkmark$ & 33.1 & $\text{10.3}/\text{4.4}$ & 32.0 & $\text{9.6}/\text{4.1}$ \\
{TwoStream}~\cite{chen2022two} & S3D & $\checkmark$ & $\checkmark^{*}$ & \baseline{25.4} & -/- & \baseline{25.3} & -/- \\
TwoStream~\cite{chen2022two} & S3D &  & $\checkmark^{*}$ & 28.9 & -/- & 28.5 & -/- \\
BN-TIN~\cite{zhou2021improving} & GoogLeNet & & $\checkmark$ & 33.6 & $\text{13.9}/\text{3.4}$ & 33.1 & $\text{13.5}/\text{3.0}$ \\    
CorrNet~\cite{hu2023continuous} & ResNet18 & & $\checkmark$ & 30.6 & -/- & 30.1 & -/- \\ 
Contrastive~\cite{gan23contrastive} & ResNet18 & & & 26.0 & $\text{11.5}/\text{3.0}$ & 25.3 & $\text{11.2}/\text{3.5}$ \\
\hline
CoSign~\cite{jiao2023cosign} & ST-GCN & $\checkmark$ & & 28.1 & {-/-} & 27.2 & {-/-} \\
\baseline{MultiSignGraph} & GCN & & $\checkmark$ & 27.3 & $\text{7.9}/\text{2.3}$ & 26.4 & $\text{7.8}/\text{2.1}$ \\
\baseline{MixSinGraph} & GCN & & $\checkmark$ & \baseline{25.1} & $\text{6.4}/\text{2.1}$ & \baseline{25.0} & $\text{7.0}/\text{1.6}$ \\
\end{tabular}}  
\caption{Comparison of CSLR performance on CSL-daily dataset.}
\label{tab:CSLdailyCSLR} 
\end{table}

 \subsection{Visualization of MixSignGraph}
To verify whether our model can effectively capture sign-related features, we select a sign video from the PHOENIX14T test set and visualize the constructed graph structure in both stages. 
In Figure~\ref{fig:graphbuilding}, we show the graphs in the $LSG$, $TSG$ and $HSG$ modules. For clarity, only a subset of the edges is displayed. 
In the $LSG$ module, our model links nodes with similar contents or related semantic representation (\eg hand regions and face regions), to extract better intra-frame cross-region features. 
In the $TSG$ module, our model builds edges among adjacent frames to track dynamic changes in gestures and facial expressions, to capture inter-frame cross-region features.  
In the $HSG$ module, our model connects the same regions with different granularities (\ie the corresponding regions in feature maps), to enhance one-region features. 
In addition, we also highlight some `unimportant' edges between nodes in the background in red. As seen in Figure~\ref{fig:graphbuilding}, the background nodes in the $LSG$ module are naturally connected to their neighboring nodes, and there are still a few background nodes connected in the $TSG$ module. Fortunately, background nodes do not `disturb' nodes in sign-related regions, demonstrating the effectiveness of our model.

\begin{table*}[t]
	\centering 
	\resizebox{0.95\textwidth}{!}{
\begin{tabular}{l|cccc|ccccc|ccccc} 
\multirow{3}*{Sign2Gloss2Text}&\multicolumn{4}{c|}{Extra cues}& \multicolumn{10}{c}{PHOENIX14T}   \\
&&&&&\multicolumn{5}{c|}{DEV} & \multicolumn{5}{c}{TEST}\\ 
\cline{2-15}
& {F/M}&{H}&{S}&{P}&ROUGE &BLEU1& BLEU2& BLEU3& BLEU4& ROUGE& BLEU1& BLEU2& BLEU3& BLEU4 \\
\shline
SL-Luong~\cite{cihan2018neural}&&&&\checkmark&44.14&42.88&30.30&23.02&18.40&43.80&43.29&30.39&22.82&18.13\\
Joint-SLRT~\cite{camgoz2020sign} &&&& \checkmark&47.73&34.82&27.11&22.11&-&48.47&35.35&27.57&22.45\\
SignBT~\cite{zhou2021improving} &&&& \checkmark&49.53&49.33&36.43&28.66&23.51&49.35&48.55&36.13&28.47&23.51\\
STMC-Transf~\cite{yin2020better}&\checkmark&\checkmark&\checkmark&\checkmark&46.31&48.27&35.20&27.47&22.47&46.77&48.73&36.53&29.03&24.00\\
MMTLB~\cite{chen2022simple}&&&&\checkmark&50.23&50.36&37.50&29.69&24.63&49.59&49.94&37.28&29.67&24.60\\
RTG-Net~\cite{gan2023towards}&\checkmark&\checkmark&\checkmark&\checkmark&50.18&{51.17}&{37.95}&{29.88}&{25.95}&{50.04}&{50.87}&{37.95}&{29.74}&{25.87}\\
TwoStream-SLT~\cite{chen2022two}&\checkmark&\checkmark&\checkmark&\checkmark&52.01&\better{52.35}&\better{39.76}&\better{31.85}&\better{26.47}&\better{51.59}&\better{52.11}&\better{39.81}&\better{32.00}&\better{26.71}\\
\better{MultiSignGraph} &&&& \checkmark&\better{52.13}&{51.96}&{39.37}&{31.68}&{26.42}& 50.21 &50.76&38.23&30.26&24.91\\ 

\better{MixSignGraph}&&&& \checkmark&\better{52.59}&\better{52.40}&\better{39.84}&\better{ 31.95}&\better{26.56}& \better{51.46} &\better{52.35}&\better{39.23}&\better{32.26}&\better{26.04}\\ 
\toprule 
Sign2Text& {F/M}&{H}&{S}&{P}&ROUGE &BLEU1& BLEU2& BLEU3& BLEU4& ROUGE& BLEU1& BLEU2& BLEU3& BLEU4 \\
\toprule
Joint-SLRT~\cite{camgoz2020sign} &&&& \checkmark&-&47.26&34.40&27.05&22.38&-&46.61&33.73&26.19&21.32\\
HST-GNN&\checkmark&\checkmark&\checkmark&&-&46.10&33.40&27.50&22.6&-&45.20&34.70&27.50&22.60\\
STMC-T~\cite{zhou2021spatial}&\checkmark&\checkmark&\checkmark&\checkmark&48.24&47.60&36.43&29.18&24.09&46.65&46.98&36.09&28.70&23.65\\
SignBT~\cite{zhou2021improving}&&&& \checkmark&50.29&51.11&37.90&29.80&24.45&49.54&50.80&37.75&29.72&24.32\\
CoSLRT~\cite{gan23contrastive}&&&&&{52.47}&{52.29} &{39.60}&{31.34}&{27.83} &{52.24} &{52.48}&{41.17}&{32.30}&{27.88}\\
MMTLB~\cite{chen2022simple}&&&&\checkmark&53.10&53.95&41.12&33.14&27.61&52.65&53.97&41.75&33.84&28.39\\
TwoStream-SLT~\cite{chen2022two}&\checkmark&\checkmark&\checkmark&\checkmark&54.08&54.32&41.99&34.15&28.66&53.48&54.90&42.43&34.46&28.95\\

\better{MultiSignGraph} &&&& \checkmark&\better{55.23}&\better{54.94}&\better{42.38}& \better{34.39}&\better{28.95}&\better{53.63}&\better{54.18}&\better{41.71}&\better{33.73}&\better{28.26}\\
\better{MixSignGraph} &&&& \checkmark&\better{55.77}&\better{55.01}&\better{42.64}& \better{34.94}&\better{29.00}&\better{53.84}&\better{54.90}&\better{42.53}&\better{34.50}&\better{28.97}\\ 
\toprule
GlossFree-Sign2Text& {F/M}&{H}&{S}&{P}&ROUGE &BLEU1& BLEU2& BLEU3& BLEU4& ROUGE& BLEU1& BLEU2& BLEU3& BLEU4 \\
\toprule
SL-Luong~\cite{cihan2018neural}&&&&\checkmark&31.80&31.87&19.11&13.16&9.94&31.80&32.24&19.03&12.83&9.58\\
TSPNet~\cite{li2020tspnet} & & & &$\checkmark$&-&-&-&-&-&34.96&36.10&23.12&16.88&13.41\\
GASLT~\cite{yin2023gloss}&&&&&-&-&-&-&-&39.86&39.07&26.74&21.86&15.74\\
CSGCR~\cite{zhao2021conditional}&&&&\checkmark&38.96&35.85&24.77&18.65&15.08&38.85&36.71&25.40&18.86&15.18\\
GFSLT~\cite{yin2023gloss}&&&&\checkmark&40.93&41.97&31.04&24.30&19.84&40.70&41.39&31.00&24.20&19.66\\
GFSLT-VLP~\cite{yin2023gloss}&&&&\checkmark&43.72&44.08&33.56&26.74&22.12&42.49&43.71&33.18&26.11&21.44\\
\baseline{MultiSignGraph} &&&& \checkmark&\baseline{51.17} &\baseline{50.57}&\baseline{37.77}  &\baseline{29.86}&\baseline{24.62}&\baseline{50.11} &\baseline{49.63}&\baseline{37.18}  &\baseline{29.25}&\baseline{23.94}\\ 
\baseline{MixSignGraph} &&&& \checkmark&\baseline{51.71} &\baseline{51.07}&\baseline{37.97}  &\baseline{29.98}&\baseline{24.87}&\baseline{51.14} &\baseline{50.01}&\baseline{38.04}  &\baseline{29.95}&\baseline{24.02}\\ 
\end{tabular}}  
\caption{Comparison of SLT performance on PHOENXI14T dataset.} 
\label{tab:SLTPhoenix14T} 
\end{table*}

\begin{table*}[t]
	\centering 
	\resizebox{0.95\textwidth}{!}{
\begin{tabular}{l|cccc|ccccc|ccccc} 
\multirow{3}*{Sign2Gloss2Text}&\multicolumn{4}{c|}{Extra cues}& \multicolumn{10}{c}{CSL-Daily }   \\
&&&&&\multicolumn{5}{c|}{DEV} & \multicolumn{5}{c}{TEST}\\ 
\cline{2-15}
& {F/M}&{H}&{S}&{P}&ROUGE &BLEU1& BLEU2& BLEU3& BLEU4& ROUGE& BLEU1& BLEU2& BLEU3& BLEU4 \\
\shline
SL-Luong~\cite{cihan2018neural}&&&&\checkmark&40.18&41.46&25.71&16.57&11.06&40.05&41.55&25.73&16.54&11.03\\
Joint-SLRT~\cite{camgoz2020sign}&&&&\checkmark&44.18&46.82&32.22&22.49&15.94&44.81&47.09&32.49&22.61&16.24\\
SignBT~\cite{zhou2021improving}&&&&\checkmark&48.38&50.97&36.16&26.26&19.53&48.21&50.68&36.00&26.20&19.67\\
MMTLB~\cite{chen2022simple}&&&&\checkmark&51.35&50.89&37.96&28.53&21.88&51.43&50.33&37.44&28.08&21.46\\
TwoStream-SLT~\cite{chen2022two}&\checkmark&\checkmark&\checkmark&\checkmark&\better{53.91}&\better{53.58}&\better{40.49}&\better{30.67}&\better{23.71}&\better{54.92}&\better{54.08}&\better{41.02}&\better{31.18}&\better{24.13}\\
\better{MultiSignGraph} &&&&\checkmark &52.16 &53.25 &40.46 &30.12 &23.14&52.26 &52.76 &40.78 &30.39 &23.06\\ 
\better{MixSignGraph} &&&&\checkmark &\better{53.48} &\better{53.73} &\better{41.53}&\better{30.70}&\better{23.76} &\better{53.65} &\better{54.07} &\better{41.04}&\better{31.32}&\better{24.44}\\ 
\toprule
Sign2Text& {F/M}&{H}&{S}&{P}&ROUGE &BLEU1& BLEU2& BLEU3& BLEU4& ROUGE& BLEU1& BLEU2& BLEU3& BLEU4 \\
\toprule
Joint-SLRT~\cite{camgoz2020sign}&&&&\checkmark&37.06&37.47&24.67&16.86&11.88&36.74&37.38&24.36&16.55&11.79\\
SignBT~\cite{zhou2021improving}&&&&\checkmark&49.49&51.46&37.23&27.51&20.80&49.31&51.42&37.26&27.76&21.34\\
Contrastive~\cite{gan23contrastive}&&&&&50.34&51.97&37.10&27.53&21.79&50.73&52.31&37.37&27.89&21.81\\
MMTLB~\cite{chen2022simple}&&&&\checkmark&53.38&53.81&40.84&31.29&24.42&53.25&53.31&40.41&30.87&23.92\\
TwoStream-SLT~\cite{chen2022two}&\checkmark&\checkmark&\checkmark&\checkmark&\better{55.10}&\better{55.21}&\better{42.31}&\better{32.71}&\better{25.76}&\better{55.72}&\better{55.44}&\better{42.59}&\better{32.87}&\better{25.79}\\

\better{MultiSignGraph}&&&&\checkmark&53.49&54.93&41.73&32.30&25.56&53.65 & 54.67&41.60&32.33&25.78\\ 
\better{MixSignGraph}&&&&\checkmark&\better{54.54}&\better{55.87}&\better{42.45}&\better{32.75}&\better{25.77}&\better{54.67}&\better{ 55.41}&\better{42.43}&\better{32.84}&\better{25.87}\\  
\toprule
Gloss-free SLT& {F/M}&{H}&{S}&{P}&ROUGE &BLEU1& BLEU2& BLEU3& BLEU4& ROUGE& BLEU1& BLEU2& BLEU3& BLEU4 \\
\toprule
SL-Luong~\cite{cihan2018neural}&&&&\checkmark&34.28&34.22&19.72&12.24&7.96&34.54&34.16&19.57&11.84&7.56\\
GASLT~\cite{yin2023gloss}&&&&\checkmark&-&-&-&-&&20.35&19.90&9.94&5.98&4.07\\
GFSLT-VLP~\cite{zhou2023gloss}&&&&\checkmark&36.44&39.20&25.02&16.35&11.07&36.70&39.37&24.93&16.26&11.00\\ 
Sign2GPT~\cite{wong2024sign2gpt}&&&&\checkmark*&  -&-&-&-&-&42.36&41.75& 28.73& 20.60 &15.40 \\
\better{MultiSignGraph} &&&&\checkmark&\better{48.51}&\better{49.72}&\better{35.99}&\better{26.53}&\better{19.97}&\better{49.11}&\better{49.97}&\better{36.54}&\better{27.10}&\better{20.59}\\ 
\better{MixSignGraph}&&&&\checkmark&\better{49.16}&\better{49.98}&\better{36.42}&\better{26.89}&\better{20.43}&\better{49.93}&\better{50.24}&\better{36.91}&\better{27.54}&\better{20.78}\\ \\ 
	\end{tabular}} 
	\caption{Comparison of SLT performance on CSL-Daily dataset. $\checkmark^*$: using DINOv2 model pre-trained on LVD-142M image dataset~\cite{oquab2024dinov2} and GPT2 model pre-trained on WebText~\cite{radford2019language}.} 
	\label{tab:SLTCSL-Daily} 
\end{table*}

\begin{table*}[t]
	\centering 
	\resizebox{0.95\textwidth}{!}{
\begin{tabular}{l|cccc|ccccc|ccccc} 
\multirow{3}*{Gloss-free SLT}&\multicolumn{4}{c|}{Extra cues}& \multicolumn{10}{c}{How2Sign}   \\
&&&&&\multicolumn{5}{c|}{DEV} & \multicolumn{5}{c}{TEST}\\ 
\cline{2-15}
& {F/M}&{H}&{S}&{P}&ROUGE &BLEU1& BLEU2& BLEU3& BLEU4& ROUGE& BLEU1& BLEU2& BLEU3& BLEU4 \\ 
\toprule
SLT-IV~\cite{tarres2023sign}&&&&\checkmark&-&35.20&20.62&13.25&8.89&-& 34.01&19.30&12.18&8.03\\  
\better{Ours} &&&& \checkmark& \better{29.24} &\better{34.82} &\better{22.47} &\better{15.61} &\better{11.28}&\better{28.01}&\better{34.74}&\better{20.83}&\better{14.41}&\better{10.41}\\ 
\end{tabular}} 
\caption{Comparison of SLT performance on How2Sign dataset.} 
	\label{tab:How2Sign}
\end{table*}

\begin{table*}[t]
	\centering 
	\resizebox{0.95\textwidth}{!}{
\begin{tabular}{l|cccc|ccccc|ccccc} 
\multirow{3}*{Gloss-free SLT}&\multicolumn{4}{c|}{Extra cues}& \multicolumn{10}{c}{OpenASL}   \\
&&&&&\multicolumn{5}{c|}{DEV} & \multicolumn{5}{c}{TEST}\\ 
\cline{2-15}
& {F/M}&{H}&{S}&{P}&ROUGE &BLEU1& BLEU2& BLEU3& BLEU4& ROUGE& BLEU1& BLEU2& BLEU3& BLEU4 \\ 
\toprule 
Conv-GRU\cite{shi2022open}&&&&\checkmark&16.25&16.72&8.95&6.31&4.82&16.10&16.11&8.85&6.18&4.58\\
I3D-transformer\cite{shi2022open}&&&&\checkmark& 18.88&18.26&10.26&7.17&5.60 &18.64&18.31&10.15&7.19& 5.66\\
Open\cite{shi2022open}&\checkmark&\checkmark&&\checkmark&20.43&20.10&11.81&8.43&6.57&21.02&20.92&12.08&8.59&6.72\\
\toprule 
\better{Ours} &&&&\checkmark&\better{25.41}& \better{26.82}& \better{16.70}& \better{11.48}& \better{8.36}& \better{25.71} &\better{26.65}&\better{16.55}&\better{11.68}&\better{8.69}\\ 
\end{tabular}} 
\caption{Comparison of SLT performance on OpenASL dataset.} 
\label{tab:OpenASL}
\end{table*}

\begin{table*}[t]
\tablestyle{1pt}{1.0}
\centering \resizebox{0.99\textwidth}{!}{
\begin{tabular}{*l|c|c|c|c|c|c}
{Model} &Task &{Visual backbone}&{Translation model} & FLOPs (G)& Parameters(M) &Pre-trained Dataset\\ 
\shline 
CrossL-Two&CSLR&Dual S3D&-& - &-& CSL-Daily+Phoenix14+Phoenix14T\\  
{SIGN2GPT}&Gloss-free SLT &Dino-V2 ViT&GPT2&-&1771.65&   LVD-142M+WebText\\
 TwoStream&CSLR+S2T+S2G2T&Dual S3D&mBART    &323.41 & 405.38 & Kinetics-400+CC25\\     
{MultiSigngraph}&CSLR+S2T+S2G2T+gloss-free SLT &SignGraph&mBART   &256.13&395.80& ImageNet-1k+CC25\\ 
{MixSigngraph}&CSLR+S2T+S2G2T+gloss-free SLT  &SignGraph&mBART  &320.88&402.20&ImageNet-1k+CC25\\
\end{tabular}}   
		\caption{Model details.}
		\label{tab:flops}  
\end{table*} 

\subsection{Qualitative Results}
\subsubsection{CSLR Results} As shown in Table~\ref{tab:SLRQualitative}, we conduct qualitative analysis for our MultiSignGraph (\ie the model in conference paper) and MixSignGraph in the CSLR task, and show one sample from the test set of PHOENIX14, PHOENIX14T and CSL-Daily, respectively.
It can be found that MixSignGraph yields more accurate gloss predictions than MultiSignGraph, demonstrating that the newly-proposed MixSignGraph is more effective. 

\subsubsection{SLT Results} 
As shown in Table~\ref{tab:SLTQualitative}, we present qualitative analysis of our MixSignGraph in the SLT task, and show one sample from the test set of PHOENIX14T, CSL-Daily, How2Sign and OpenASL, respectively.   
In can be found that the gloss-based Sign2Text model achieves the best performance while our MixSignGraph pre-trained with TCP also achieves satisfactory translation results.  

\subsection{Comparisons on CSLR tasks} 
\subsubsection{Evaluation on PHOENIX14T dataset} 
As shown in Table~\ref{tab:Phoenix}, we compare our model with existing models on CSLR performance, and we provide both the performances on the validation set (\ie `DEV') and test set (\ie `TEST'). 
Most of the current models adopt existing CNN-based backbones, and achieve convincing performances by injecting extra cues~\cite{zhou2020spatial, chen2022two, zuo2022c2slr}, adding extra constraints~\cite{gan23contrastive,min2021visual,hao2021self} or introducing attention mechanism~\cite{hu2023continuous}. 
The GCN-based model CoSign~\cite{jiao2023cosign} mainly relies on pre-processed fine-grained skeleton data, and achieves 19.5\%, 20.1\% WER on dev, test set. While HST-GCN~\cite{kan2022sign} adopts both CNN-based backbone and GCN-based backbone to extract RGB features and skeleton features respectively, and achieves 19.5\%, 19.8\% WER on dev, test set. 
It is worth mentioning that the SOTA model CrossL-Two~\cite{wei2023improving} utilizes both RGB features and fine-grained skeleton features (\ie keypoints in hands, body and face), and pre-trains its backbone on both PHOENIX14 and CSL-daily datasets, achieving 16.9\%, 18.5\% WER on dev, test set respectively. 
When moving to our MixSignGraph, it does not use any extra cues or pre-train on other SL datasets, but it still achieves comparable performance with CrossL-Two on test set. While on dev set, our model even outperforms CrossL-Two model by 0.2\% WER.

\subsubsection{Evaluation on PHOENIX14 dataset} We conduct a comparative analysis of our model and current CSLR models using the PHOENIX14 dataset. 
As illustrated in Table~\ref{tab:Phoenix}, our straightforward yet robust MixSignGraph model achieves the superior performance (16.5\% WER) on the dev set and surpasses the majority of existing models, thus demonstrating the effectiveness of our MixSignGraph.

\subsubsection{Evaluation on CSL-daily dataset} 
We also compare our model with other methods on CLSR performance, by using the CSL-daily dataset.  
Table~\ref{tab:CSLdailyCSLR} shows that our model achieves 25.1\%, 25.0\% WER on the dev, test set, respectively. It indicates that our model can surpass the SOTA model by only using RGB modality.

\subsection{Comparisons on SLT tasks} 

\subsubsection{Evaluation on Phoenix14T dataset}
As shown in Table~\ref{tab:SLTPhoenix14T}, we compare our model with existing models using PHOENIX14T dataset on SLT tasks, which include Sign2Gloss2Text, Sign2Text, and Gloss-free Sign2Text (also known as Gloss-free SLT).   
Our previous model MultiSignGraph achieves comparable performance to the SOTA model (TwoStream) on both Sign2Gloss2Text and Sign2Text tasks, while our newly-proposed model MixSignGraph outperforms the SOTA model on both Dev and Test sets. 
When moving to the gloss-free SLT task, our MixSignGraph pre-trained with TCP mechanism significantly surpasses the current SOTA model, \eg 8.49 ROUGE score higher than GFSLT-VLP~\cite{yin2023gloss}. Besides, MixSignGraph also narrows the performance gap between the gloss-free SLT model and the gloss-based SLT model, \ie 52.27 ROUGE score vs. 55.57 ROUGE score. 

\subsubsection{Evaluation on CSL-daily dataset}
As shown in Table~\ref{tab:SLTCSL-Daily}, we also show the SLT performance of our MixSignGraph on CSL-daily dataset. It can be found that MixSignGraph outperforms the SOTA model on all the Sign2Gloss2Text, Sign2Text and gloss-free SLT tasks, especially on the gloss-free SLT task (\ie 12.72 ROUGE score higher than the SOTA model).

\subsubsection{Evaluation on How2Sign dataset} How2Sign is a large American Sign Language dataset, and currently, there are only a few papers that attempt to provide baseline models trained on this dataset. 
Considering that How2Sign dataset only provides sentence labels (\ie NO gloss labels), we compare our proposed model with existing models in terms of gloss-free SLT performance.
As shown in Table~\ref{tab:How2Sign}, our MixSignGraph pre-trained with TCP achieves a BLEU4 score of 11.28/10.41 on dev/test set and outperforms the baseline model, thereby provides a new baseline for future work.

\subsubsection{Evaluation on OpenASL dataset} Similar to How2Sign dataset, OpenASL is also a large dataset that contains only sentence labels and has a very rich vocabulary. 
There is only one paper~\cite{shi2022open} that attempts to provide baseline models trained on this dataset. 
We compare our proposed model with these baseline models in terms of gloss-free SLT performance.
As shown in Table \ref{tab:OpenASL}, {our model} achieves the SOTA performance and provides a new baseline for future work.

\subsection{Model Comparison and Analysis}
To provide a more comprehensive analysis, we compare our model with the existing models in the following aspects: tasks, architecture information, flops, number of parameters, and pre-trained datasets. 
As shown in Table~\ref{tab:flops}, among the existing models, the TwoStream model adopts a dual S3D backbone and pre-train their backbone with Kinetics-400, which is an action recognition dataset. 
Based on the backbone in TwoStream, the CrossL-Two model further pre-trains their backbone with existing SL datasets. 
SIGN2GPT adopts Dino-V2 ViT as visual backbone which is pre-trained on LVD-142M (a very large vision dataset), and adopts GPT2 as the translation model which is also pre-trained on a very large text dataset WebText.
In regard to our model, it contains fewer parameters and FLOPs, and it is only pre-trained on small datasets (\ie ImageNet-1K and CC25 datasets), while achieving the SOTA performance on multiple SL tasks.

\section{Conclusion}
In this paper, we propose a simple yet effective MixSignGraph architecture, which represents a sign sequence as graph structures to extract sign-related features at the graph level. 
Specifically, to learn the correlation of cross-region features in one frame, we propose a Local Sign Graph $LSG$ module, which dynamically builds edges between nodes in one frame and aggregates intra-frame cross-region features. To track the interaction of cross-region features among adjacent frames, we propose a Temporal Sign Graph $TSG$ module, which dynamically builds edges between nodes among adjacent frames and captures inter-frame cross-region features. 
To combine coarse-grained and fine-grained features in one region at the same time, we design a hierarchical sign graph $HSG$ module, which connects the same regions in different-granularity feature maps and allows for bidirectional information exchange between feature maps.
Extensive experimental results on five sign language datasets and common sign language tasks (\ie both CSLR and SLT) show that our model achieves a good performance without relying on any additional cues.

\section{Limitations and Borader Impacts} 
Although the proposed MixSignGraph model can achieve promising results only using RGB modality, there are still many limitations of MixSignGraph. Here, we list some potential ideas that can be further explored to improve performance of SL tasks. 
First, our model introduces {KNN}-based graph building method, in which hyperparameter $K$ (\ie number of edges) may affect model performance. Therefore, a better way to build graphs may bring more effective cross-region features.
Second, our TCP mechanism for the gloss-free SLT task works well based on processed text, but it also encounters problems such as difficulty in CTC loss convergence, especially in cases with large vocabularies (\eg OpenASL). Therefore, more effective methods to obtain pseudo labels for CTC pre-training are expected. 
In addition, we also highlight the potential negative social impacts.
First, deep learning thrives on large-scale training data, thus may lead to the large training cost. 
Second, our method may experience unpredictable failures, so it should not be used in scenarios where such failures could have serious consequences.
Third, our method is a data-driven approach and its performance may be influenced by biases in data, thus caution is advised in the data collection process.

\bibliographystyle{IEEEtran}
\bibliography{SignGraph2}
\begin{IEEEbiography}[{\includegraphics[width=1.0in,height=1.25in,clip,keepaspectratio]{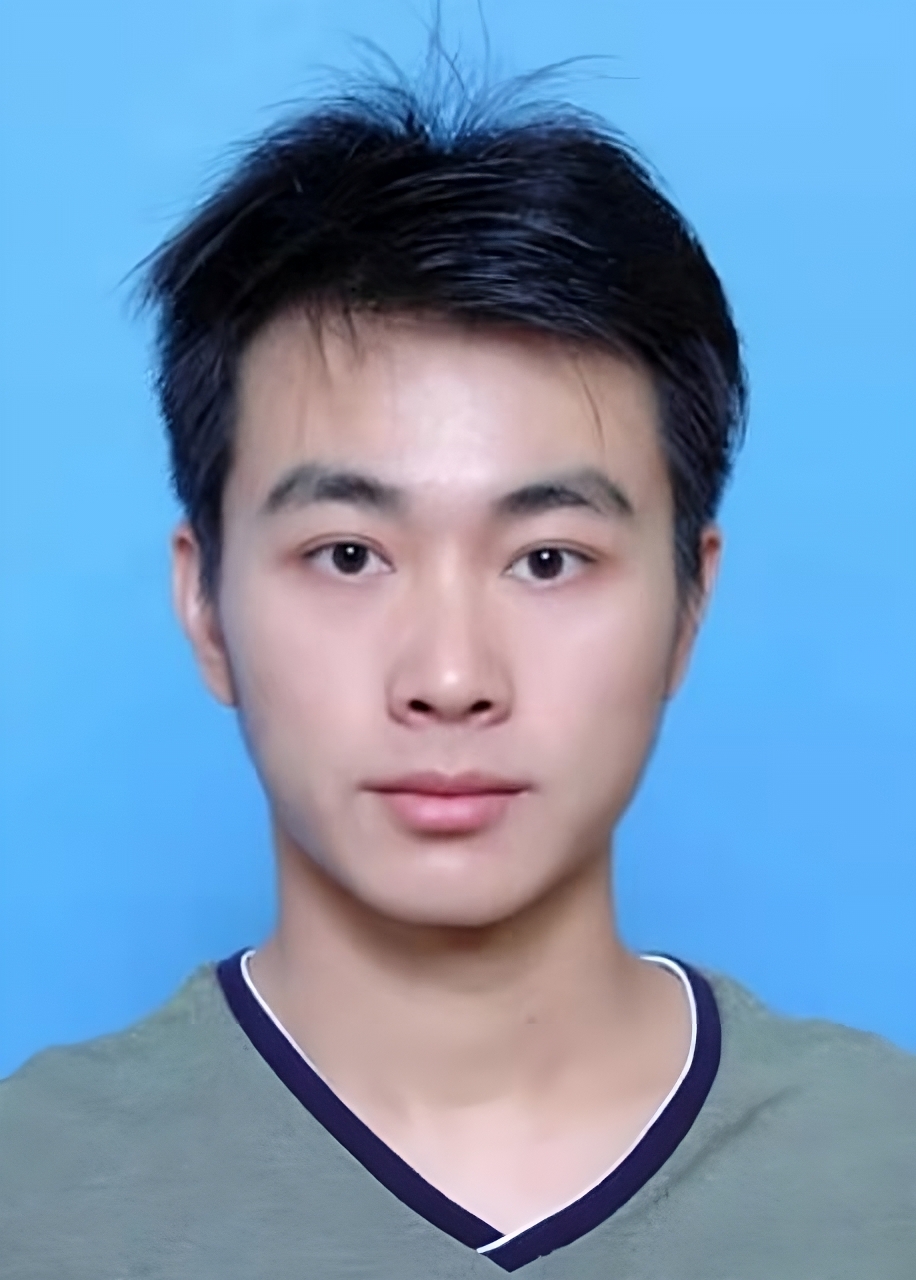}}] {Shiwei Gan} (Student Member, IEEE) received his B.E.
degree in computer science from Hunan
University, Changsha, in 2019. He
is currently a Ph.D. candidate in the
Department of Computer Science at
Nanjing University, Nanjing. His
research interests include sign language
processing, multimodal large models, and large language models.
\end{IEEEbiography} 
\begin{IEEEbiography}[{\includegraphics[width=1in,height=1.25in,clip,keepaspectratio]{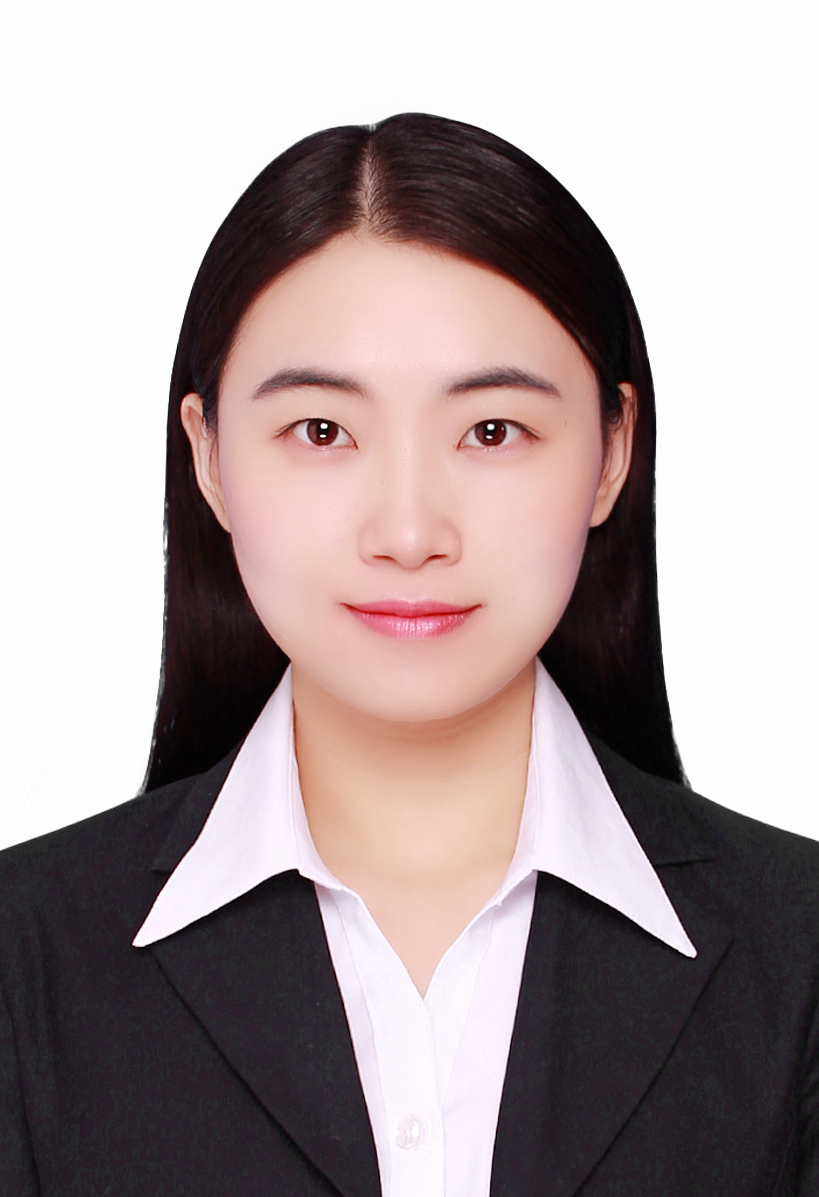}}] {Yafeng Yin} (Member, IEEE) received her Ph.D. degree in computer science from Nanjing University, China in 2017. She is currently an assistant professor in the School of Intelligent Software and Engineering at Nanjing University. Her research interests include sign language processing, human action recognition, deep learning, etc. She has published over 40 papers in ACM Transactions on Information Systems, IEEE Transactions on Computers, IEEE Transactions on Mobile Computing, CVPR, IJCAI, ACM MM, ACM UbiComp, IEEE INFOCOM, etc.
\end{IEEEbiography}
\vspace{-4mm}
\begin{IEEEbiography}[{\includegraphics[width=1in,height=1.25in,clip,keepaspectratio]{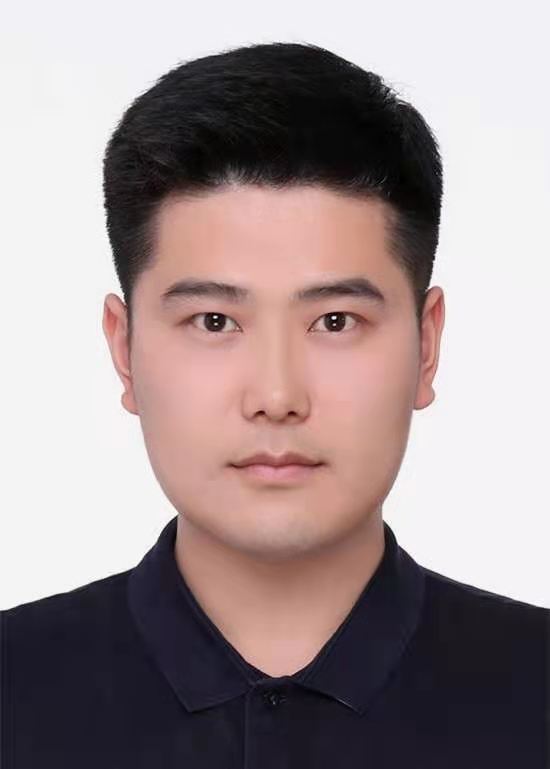}}] {Zhiwei Jiang} (Member, IEEE) received his Ph.D. degree in computer science from Nanjing University, China in 2018.
He is currently an assistant professor in the School
of Intelligent Software and Engineering at Nanjing
University. His research interests include natural
language processing, machine learning, etc.
\end{IEEEbiography} 
\begin{IEEEbiography}[{\includegraphics[width=1in,height=1.25in,clip,keepaspectratio]{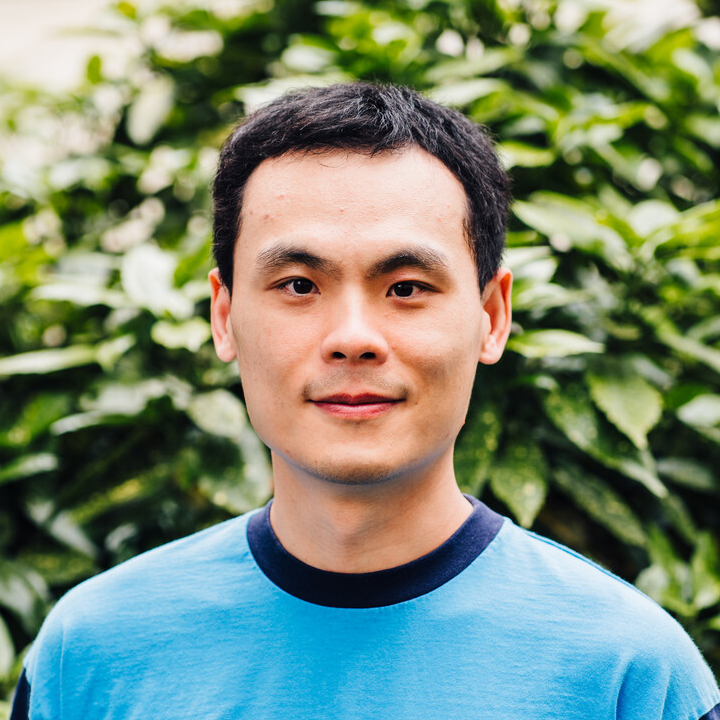}}] {Hongkai Wen} (Senior Member, IEEE) received the DPhil
degree from the University of Oxford, and became
a post-doctoral researcher in a joint project between Oxford Computer Science and Robotics Institute. He is a Professor with the Department of Computer Science, University of Warwick. Broadly speaking, his research belongs to the area of CyberPhysical Systems, which use networked smart devices to sense and interactive with the physical world.
\end{IEEEbiography}

\balance

\begin{IEEEbiography}[{\includegraphics[width=1in,height=1.25in,clip,keepaspectratio]{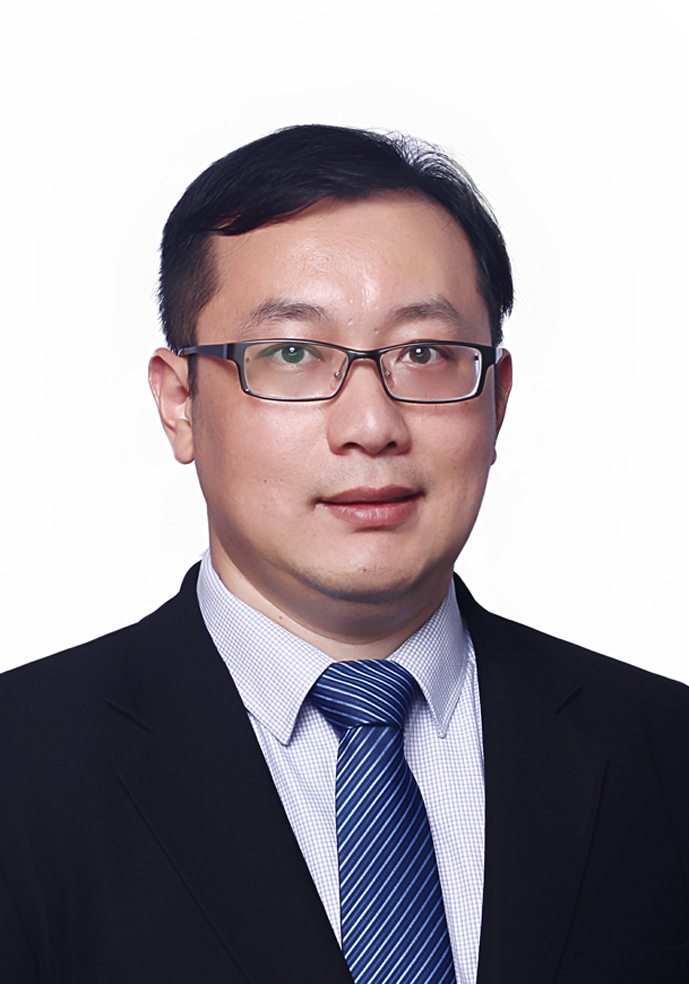}}] {Lei Xie} (Senior Member, IEEE) received his B.S. and Ph.D. degrees from
Nanjing University, China in 2004 and 2010, respectively, all in computer science. He is currently a
professor in the Department of Computer Science
and Technology at Nanjing University. His research interests include edge computing, pervasive computing, machine learning, etc. 
He has published over 100 papers in ACM/IEEE Transactions on
Networking, IEEE Transactions on Mobile Computing, IEEE Transactions on Parallel and Distributed
Systems, CVPR, IJCAI, ACM MM, ACM MobiCom, ACM UbiComp, IEEE INFOCOM, etc.
\end{IEEEbiography} 

\begin{IEEEbiography}[{\includegraphics[width=1in,height=1.25in,clip,keepaspectratio]{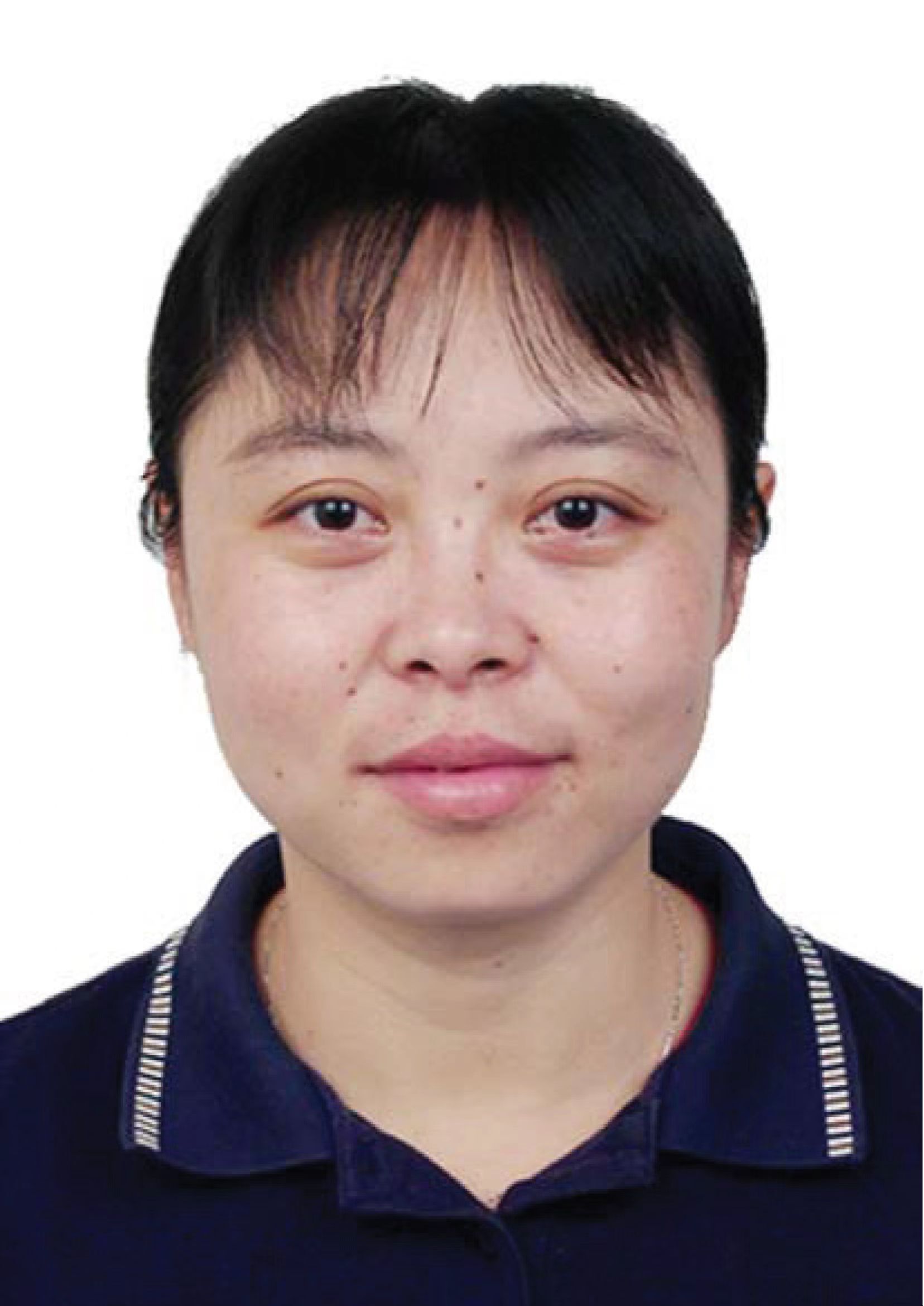}}] {Sanglu Lu}  (Member, IEEE) received the BS, MS, and PhD degrees from Nanjing University, China, in 1992, 1995, and 1997, respectively, all in computer science. She is currently a professor in the Department of Computer Science and Technology at Nanjing University. Her research interests include machine learning, pervasive computing, etc. 
\end{IEEEbiography}

\end{document}